\documentclass[twoside,twocolumn,english,british]{article}
\usepackage[T1]{fontenc}
\usepackage[utf8]{inputenc}
\usepackage[a4paper]{geometry}
\usepackage{fancyhdr}
\usepackage{array}
\usepackage{rotating}
\usepackage{multirow}
\usepackage{amsmath}
\usepackage{amssymb}
\usepackage{graphicx}
\usepackage{esint}

\makeatletter

\DeclareFontEncoding{LGR}{}{}
\DeclareTextSymbol{\~}{LGR}{126}
\newcommand{\lyxmathsym}[1]{\ifmmode\begingroup\def\b@ld{bold}
  \text{\ifx\math@version\b@ld\bfseries\fi#1}\endgroup\else#1\fi}

\providecommand{\tabularnewline}{\\}

\DeclareUnicodeCharacter{0391}{\ensuremath{\Alpha}}
\DeclareUnicodeCharacter{03B1}{\ensuremath{\alpha}}
\DeclareUnicodeCharacter{0392}{\ensuremath{\Beta}}
\DeclareUnicodeCharacter{03B2}{\ensuremath{\beta}}
\DeclareUnicodeCharacter{0393}{\ensuremath{\Gamma}}
\DeclareUnicodeCharacter{03B3}{\ensuremath{\gamma}}
\DeclareUnicodeCharacter{0394}{\ensuremath{\Delta}}
\DeclareUnicodeCharacter{03B4}{\ensuremath{\delta}}
\DeclareUnicodeCharacter{0395}{\ensuremath{\Epsilon}}
\DeclareUnicodeCharacter{03B5}{\ensuremath{\epsilon}}
\DeclareUnicodeCharacter{0396}{\ensuremath{\Zeta}}
\DeclareUnicodeCharacter{03B6}{\ensuremath{\zeta}}
\DeclareUnicodeCharacter{0397}{\ensuremath{\Eta}}
\DeclareUnicodeCharacter{03B7}{\ensuremath{\eta}}
\DeclareUnicodeCharacter{0398}{\ensuremath{\Theta}}
\DeclareUnicodeCharacter{03B8}{\ensuremath{\theta}}
\DeclareUnicodeCharacter{0399}{\ensuremath{\Iota}}
\DeclareUnicodeCharacter{03B9}{\ensuremath{\iota}}
\DeclareUnicodeCharacter{039A}{\ensuremath{\Kappa}}
\DeclareUnicodeCharacter{03BA}{\ensuremath{\kappa}}
\DeclareUnicodeCharacter{039B}{\ensuremath{\Lambda}}
\DeclareUnicodeCharacter{03BB}{\ensuremath{\lambda}}
\DeclareUnicodeCharacter{039C}{\ensuremath{\Mu}}
\DeclareUnicodeCharacter{03BC}{\ensuremath{\mu}}
\DeclareUnicodeCharacter{039D}{\ensuremath{\Nu}}
\DeclareUnicodeCharacter{03BD}{\ensuremath{\nu}}
\DeclareUnicodeCharacter{039E}{\ensuremath{\Xi}}
\DeclareUnicodeCharacter{03BE}{\ensuremath{\xi}}
\DeclareUnicodeCharacter{039F}{\ensuremath{\Omicron}}
\DeclareUnicodeCharacter{03BF}{\ensuremath{\omicron}}
\DeclareUnicodeCharacter{03A0}{\ensuremath{\Pi}}
\DeclareUnicodeCharacter{03C0}{\ensuremath{\pi}}
\DeclareUnicodeCharacter{03A1}{\ensuremath{\Rho}}
\DeclareUnicodeCharacter{03C1}{\ensuremath{\rho}}
\DeclareUnicodeCharacter{03A3}{\ensuremath{\Sigma}}
\DeclareUnicodeCharacter{03C3}{\ensuremath{\sigma}}
\DeclareUnicodeCharacter{03A4}{\ensuremath{\Tau}}
\DeclareUnicodeCharacter{03C4}{\ensuremath{\tau}}
\DeclareUnicodeCharacter{03A5}{\ensuremath{\Upsilon}}
\DeclareUnicodeCharacter{03C5}{\ensuremath{\upsilon}}
\DeclareUnicodeCharacter{03A6}{\ensuremath{\Phi}}
\DeclareUnicodeCharacter{03C6}{\ensuremath{\varphi}}
\DeclareUnicodeCharacter{03A7}{\ensuremath{\Chi}}
\DeclareUnicodeCharacter{03C7}{\ensuremath{\chi}}
\DeclareUnicodeCharacter{03A8}{\ensuremath{\Psi}}
\DeclareUnicodeCharacter{03C8}{\ensuremath{\psi}}
\DeclareUnicodeCharacter{03A9}{\ensuremath{\Omega}}
\DeclareUnicodeCharacter{03C9}{\ensuremath{\omega}}
\DeclareUnicodeCharacter{00A0}{~}
\DeclareUnicodeCharacter{00AC}{\ensuremath{\neg}}
\DeclareUnicodeCharacter{00B1}{\ensuremath{\pm}}
\DeclareUnicodeCharacter{00D7}{\ensuremath{\times}}
\DeclareUnicodeCharacter{00F7}{\ensuremath{\div}}
\DeclareUnicodeCharacter{2026}{\ldots}
\DeclareUnicodeCharacter{207A}{\ensuremath{^{+}}}
\DeclareUnicodeCharacter{207B}{\ensuremath{^{-}}}
\DeclareUnicodeCharacter{2020}{\ensuremath{\dagger}}
\DeclareUnicodeCharacter{2021}{\ensuremath{\ddagger}}
\DeclareUnicodeCharacter{2212}{\ensuremath{-}}
\DeclareUnicodeCharacter{2192}{\ensuremath{\rightarrow}}

\usepackage{booktabs}
\usepackage{graphicx}

\makeatother

\usepackage{babel}
\begin{document}

\title{Reconstruction of Epsilon-Machines in Predictive Frameworks and Decisional
States}

\author{Nicolas Brodu <nicolas.brodu@univ-rennes1.fr>%
\thanks{University of Rennes 1. Part of this work was done while working at
the LTSI laboratory, part while working at the CAREN laboratory.%
}}

\date{May 2011}
\maketitle
\begin{abstract}
This article introduces both a new algorithm for reconstructing epsilon-machines
from data, as well as the \emph{decisional states}. These are defined
as the internal states of a system that lead to the same decision,
based on a user-provided utility or pay-off function. The utility
function encodes some a priori knowledge external to the system, it
quantifies how bad it is to make mistakes. The intrinsic underlying
structure of the system is modeled by an epsilon-machine and its causal
states. The decisional states form a partition of the lower-level
causal states that is defined according to the higher-level user's
knowledge. In a complex systems perspective, the decisional states
are thus the {}``emerging'' patterns corresponding to the utility
function. The transitions between these decisional states correspond
to events that lead to a change of decision. The new REMAPF algorithm
estimates both the epsilon-machine and the decisional states from
data. Application examples are given for hidden model reconstruction,
cellular automata filtering, and edge detection in images.

\end{abstract}

\section{Motivation}

We are monitoring a system, and we are given a utility/cost function
for comparing predictions made about this system to what happens really.
For example, we are monitoring the weather. We have a pay-off function
$U(y,z)$ related to setting an equipment outdoor, with $y$ the weather
we predict to take our decision and $z$ what really happens. We benefit
from the equipment in the case it is outside when the weather is good,
so $U(y=\textrm{sunny},z=\textrm{sunny})=1$, while we gain nothing
when it is inside and it is raining: $U(y=\textrm{rain},z=\textrm{rain})=0$.
We miss an opportunity when we keep the equipment indoor when it could
have been useful, so $U(y=\textrm{rain},z=\textrm{sunny})=-1$ . The
equipment gets damaged under the rain, so $U(y=\textrm{sunny},z=\textrm{rain})=-2$.
We would like events telling us when to set up the equipment or not
based solely on the current system state $x$. These events are determined
by maximising the expected utility of our predictions $y$ based on
$x$.

This simple scenario is easy to transpose to more elaborated contexts.
This article presents the theoretical background for this problem,
as well as a concrete algorithm for computing the above events from
data and a utility function only. The main contributions of this article
are :
\begin{itemize}
\item A practical algorithm for computing the system states from data, which
as an intermediate step reconstructs the ε-machine \cite{CrutchfieldStatComp}.
\item The way information is encoded in the utility function, which represents
a new and clear way to represent the user knowledge independently
of the system’s intrinsic dynamics.
\item The decisional states concept, allowing a modeller to represent system
states with equivalent decisions for the user based on the preceeding
utility function.
\end{itemize}
Section \ref{sec:Related-work} relates this context to background
information and existing concepts in the litterature. Section \ref{sec:Decisional-states-setup}
introduces formally how the knowledge brought by a utility function
can be used in order to compute the internal states of the process
leading to the same decisions. Section \ref{sec:Examples} gives mathematical
examples of the theory introduced in Section \ref{sec:Decisional-states-setup}.
Section \ref{sec:Algo} details how to effectively compute decisional
and other utility-related process states from data. Section \ref{sec:Application-examples}
gives application examples, including infering hidden states from
symbolic time series, detecting patterns in cellular automata and
edges in images. A general conclusion is then given.

Free/libre source code is available and a link to the reference implementation
is given at the end of the document.

\section{\label{sec:Related-work}Background information}

\subsection{\label{sub:Causality_and_predictions}Causality, predictions and
causal states in a physical context}

\begin{figure}[h]
\begin{centering}
\includegraphics[width=1\columnwidth]{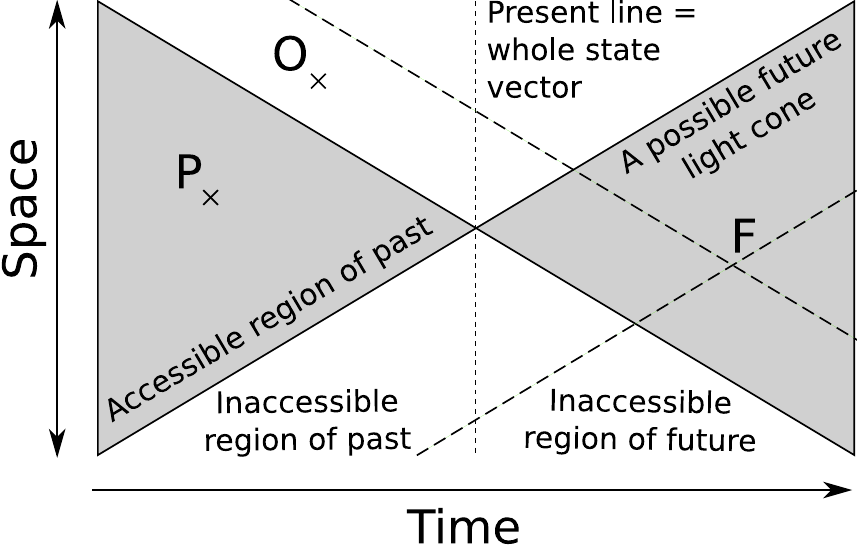}
\par\end{centering}

\caption{Light-cone representation of a prediction problem\label{fig:Light-cones}}
\end{figure}

Let us consider what a prediction means in a physical framework, where
information transfer is limited in speed. Fig. \ref{fig:Light-cones}
displays a schematic view of a system's past and possible future.

In this view the system present is a single point in state-space.
Contrast this with dynamical systems where the present is the whole
state vector, the middle vertical line in Fig. \ref{fig:Light-cones}.
Here there is no instant propagation of information, and only a small
portion of the state vector is accessible.

The past light-cone is the collection of all points that could possibly
have an a priori causal influence on the present. The future cone
is the collection of all points that might possibly be influenced
by the present state. The problem is that in order to infer correctly
the state of a point $F$ in the future cone we might potentially
need all points in the past light cone of $F$. It would theoretically
be possible to have access to the points like $P$ in the system current
past, provided in practice that we indeed recorded the value of $P$.
However there is by definition no way of getting the value of points
like $O$ that are outside the current system past light cone. Since
both points belong to the past light cone of a point $F$ in our own
future, the consequence is that even for deterministic systems we
get a statistical distribution of possible futures for a given observed
past, depending on what information present outside the current past
cone is necessary to predict the future. In other words, boundary
and/or conditions in inaccessible regions may determine part of the
future, which is well-known in physics.

Let us now consider grouping two past cones $x_{1}^{-}$ and $x_{2}^{-}$
together if they lead to the same distribution of futures $P(X^{+}|x_{1}^{-})=P(X^{+}|x_{2}^{-})$%
\footnote{In this document an upper-case $P$ is used to denote the whole distribution,
lower-case $p$ is used to denote the probability density at one point.
Similarly an upper-case $X$ denotes a whole space, whereas $x\in X$
denotes a point in that space.%
}. Suppose that point P in the past have a distinct value for pasts
$x_{1}^{-}$ and $x_{2}^{-}$. Then there is no way to recover what
the value of P was by new observations: we cannot use future knowledge
to decide between $x_{1}^{-}$ and $x_{2}^{-}$ since $P(X^{+}|x_{1}^{-})=P(X^{+}|x_{2}^{-})$.
For all practical matter these two pasts are then equivalent. Mathematically
the associated equivalence relation $x_{1}^{-}\sim x_{2}^{-}$ partitions
the system past light cones in sets $\sigma(x^{-})=\left\{ y^{-}:\, P(X^{+}|y^{-})=P(X^{+}|x^{-})\right\} $.

The sets $\sigma$ are called the \emph{causal states} of the system
\cite{ShaliziThesis}. In a discrete scenario a new observation leads
to a transition from a state $\sigma_{1}$ to a state $\sigma_{2}$.
The causal states and their transitions form a deterministic automaton:
the ε-machine \cite{CrutchfieldStatComp}. A neat result is the abstraction
of the time dependencies into the states. The transitions between
states include all dependencies from the past that could have an influence
on the future, hence the ε-machine actually forms a Markovian automaton
\cite{ShaliziThesis,DMarkovMachine}.

However unlike a traditionnal Hidden Markov Model (HMM), the ε-machine
strongly constrains the state-to-state transitions by construction.
The internal states of the ε-machine thus have a very definite meaning
(see above) and the structure of the ε-machine can be inferred from
data (see Section 5). The hidden states of the HMM are by contrast
a free parameter when training the model to best fit the data, which
needs to be estimated by techniques like maximum-likelihood or cross-validation
\cite{selectingHMMnumstates}. The differences between finite state-output
Markov processes and ε-machines is expressely highlighted in \cite{DMarkovMachine}.
The interested reader may find a comparison with alternative techniques
like Variable-Length Markov Models \cite{vlmc}, Observable Operator
Models \cite{OOM}, Predictive State Representations \cite{psr},
as well as extensions in \cite{methods_techniques_css,genNatPred,TalvitieThesis,lohr_thesis}
for example.

The causal state construction based on light-cones was introduced
in \cite{filters_ca_stat_comp}. This framework is well suited to
the analysis of physical contexts like fluid dynamics \cite{cstatesFluidDyn},
and when causal relations can naturally be traced back like in a neural
network context \cite{drrn2}. Yet the causal states construction
is not limited to light-cones. We can also cluster together data points
$x\in X$ according to the conditional distributions $P(Z|x)$ of
points $z\in Z$ in a space of predictions. The same equivalence relation
as above can be defined, except that now care must be exercised on
the interpretation: all we have defined are internal states with the
same predictive power, without referring to causality. For example,
sneezing and coughing are good predictors for being ill, though they
are the symptoms and not the cause of the illness. However, when the
space $X$ is restricted to past time (and $Z$ to future time) as
is the case in this section, it is a reasonable assumption that a
causal relation indeed provides the desired predictive ability. We
refer to the equivalence classes induced by the above relation as
causal states in this document, following the current usage of the
term, but keep in mind that prediction and causation are different
issues.

\subsection{Decisions using a utility function}

The idea of using the expected utility in order to determine which
decision to take is not new %
\footnote{The Wikipedia entry traces the history of the concept back to Bernoulli's
work in the 18th Century: http://en.wikipedia.org/wiki/Expected\_utility%
}. Usually the utility is defined as a real-valued function associated
to each outcome and quantifies the user's interest in that outcome.
Probability theory is then used to estimate the expected utility one
may get when taking different actions. The action with maximal expected
utility is then usually retained, although risk-aversion effects are
sometimes taken into account. There is an abundant litterature on
probability and decision theory \cite{DecisionTheory,BishopML,patternRec,pomdp,MDP_60}
which details these ideas.

There also exists numerous ways to combine a utility function with
a Markov model. In a (possibly Partially Observable) Markov Decision
Processes (MDP) \cite{MDP_60,pomdp}, the previous observations are
used to build a model of the world (possibly with hidden states).
Actions are then chosen according to this model in order to maximise
the expected utility that would result of that choice (including costs
and rewards). MDP consider a feasible set of actions that can be taken
at any internal state, and the effect of these actions. The utility
can be expressed as a \emph{single-parameter function of the past}
histories $U(x^{-})$ or it can be expressed in terms of a reward
$R(S=s)$ of being in state $s$ and a cost function $C(S=s,A=a)$
of taking action $a$ while in state $s$. In \cite{UtileDistinctionHMM}
each hidden state of the Hidden Markov Model is attributed a mean
and variance utility of the process to be in that state. This approach
fares well when the utility is itself a global quantity and may evolve
over time.

In the present framework, previous observations are also used to build
a model (the ε-machine). However, the ε-machine captures equivalence
classes of probability distributions of futures by definition. Hence,
all past observations within the same state will lead to the same
optimal decisions (see Section \ref{sub:Definitions}): these decisions
are taken based on predictions of what will happen next. The utility
$U(y^{+},x^{+})$ is thus here a \emph{two-parameter function of the
future}, and quantifies the benefits/costs incurred by comparing what
we thought would happen (a predicted future $y^{+}$) to what really
happens (the true future $x^{+}$). An common assumption with the
previous approaches is that the utility function encodes all the information
needed to take a decision. In the present case, when both the expected
utility and the predictions are the same we assume the user takes
the same decisions.

When dealing with utility functions based on the effect of predictions
the ε-machine naturally becomes the underlying model that is inferred
from data. In other words the utility function determines a structure
corresponding to the user knowledge on top of the ε-machine, while
the ε-machine itself represents the system's internal relations independently
of the user. This clear separation of internal structure vs. external
knowledge is a neat secondary effect of defining the utility in terms
of the effect of user predictions instead of attributing a utility
directly to each outcome.

The decisional states framework is well suited to the scenario given
in introduction, but perhaps not so well suited to reinforcement learning
\cite{UtileDistinctionsRL,RRL_overview}. However the same formalism
is applicable to any system in which a prediction {}``usefulness''
can be defined, including classical loss/utility functions like the
minimum sum of squares error between the prediction and the actual
future (See section \ref{sec:Examples}).

A framework that is closely related to reinforcement learning and
that also makes use of ε-machines has recently been proposed \cite{infoInterLearn}.
It shows that a balance between exploration and exploitation emerges
as a consequence of using the ε-machine formalism without having to
introduce that balance explicitly. That framework also relates learning
with energy minimisation, and it makes explicit the agent actions.
The present approach differs in that it introduces utility functions
and considers that all futures are not equivalent for the agent. It
would be interesting to try combining both approaches.

\subsection{Practical implementation}

A major contribution of this paper is to present a new algorithm for
reconstructing ε-machines and their extension to the decisional states
introduced in this document (See Section \ref{sec:Algo}). This REMAPF
(Reconstrution of Epsilon MAchines in a Predictive Framework) algorithm
offers more flexibility in its data representation and choice of parameters
than the previous one CSSR (Causal State Splitting Reconstruction)
\cite{CSSR}, while providing a computational performance that makes
it suitable for a large class of practical applications (See the examples
in Section \ref{sec:Application-examples}). It is possible to call
only the ε-machine reconstruction part (See Section \ref{sub:evenProcess})
of the algorithm, and thus apply it to other frameworks than the one
presently considered.

\section{\label{sec:Decisional-states-setup}Decisional states framework}

The previous section aimed at giving the reader an intuitive idea
of what the framework proposed in this document is about. This section
describes the framework formally.

\subsection{\label{sub:General-case}General problem targeted by the proposed
framework}

Let $X$ be a space comprising configurations $x$ of the system under
investigation. Let $Z$ be a space of all entities that we wish to
predict from the current system state. For example:
\begin{itemize}
\item In a symbolic series context, $X$ is the set of all past strings
up to the current point ($x\in\mathcal{A}^{*}$ with $\mathcal{A}$
the alphabet), and $Z$ is the set of all future strings after the
current point. The concrete example in Section \ref{sub:evenProcess}
highlights this case.
\item More generally for temporal systems with state space, an $x\in X$
should include all causal influence from the past that might possibly
affect the present (i.e, a past light cone, see Section \ref{sub:Causality_and_predictions}).
Similarly, $z\in Z$ is the set of future light cones. The concrete
example in Section \ref{sub:Cellular-automaton} highlights this case.
\item In the case of a non-temporal system, $X$ is defined as the relevant
space of parameters that have an influence on the system state at
the point under investigation, and similarly for $Z$ being the space
of parameters influenced by $X$. A common example in physics is the
Markov Random Field representation of lattice systems \cite{MRF_LatticeSystems}.
In an image context $X$ is the neighbourhood of a given pixel up
to a range that we assume determines the statistical distribution
of that pixel, and $Z$ is the value of the pixel \cite{AwateThesis}.
The concrete example in Section \ref{sub:Image-segmentation} highlights
this case.
\end{itemize}
For each configuration $x\in X$, we'd like to associate a prediction
$y_{x}\in Z$ amongst all possible outcomes. The actual outcome $z\in Z$
can differ from $y_{x}$: we have a range of possible $z\in Z$, and
they occur with a probability distribution $P(Z|x)$. This assumption
restricts the spaces $X$ and $Z$ we may consider, for example discrete
spaces, or continuous spaces with a canonical reference measure.

Let us now consider that the utility incurred by having acted according
to prediction $y$ when $z$ is the future that actually happens is
quantified by $U(y,z)$, independently of the particular $x$ for
which $y$ was chosen instead of $z$ (so, $U$ is a real-valued function
defined on $Z^{2}$). We could equivalently define a loss function
with $L(y,z)=-U(y,z)$. Minimising the loss is equivalent to maximising
the utility, both concepts will be used interchangeably when needed.

An important difference between the present context and typical decision-theory
frameworks (ex: \cite{MDP_60,UtileDistinctionHMM}) is thus that utility
functions have two arguments: \emph{the utility quantifies our knowledge
of how bad it is to make mistakes}. Actions are based on predictions
on what we think will happen, and are thus mapped to subsets of possible
futures (ex: {}``going out for a hike'' is mapped to {}``it won't
rain in the next hours''). Actions are implicit in the utility function:
The utility function quantifies the effect of having taken an action
based on a prediction $y$, while $z$ actually happens.

We can recover a scalar quantity at the current system state by computing
the expected utility, integrated over all possible futures that may
happen. The expected utility, in a continuous context where the integrals
exist (ex: Lebesgue measurable spaces), is:

\[
\mathbb{E}[U]=\int_{x\in X}\int_{z\in Z}U\left(y_{x},z\right)p\left(x,z\right)\mathrm{d}x\mathrm{d}z
\]

or in a discrete scenario:
\[
\mathbb{E}[U]=\sum_{x\in X}\sum_{z\in Z}U\left(y_{x},z\right)p\left(x,z\right)
\]

And for all $x\in X$ with non-null probability%
\footnote{Technically we should introduce here a set $X'=X\backslash\{x:p(x)=0\}$
of all $x\in X$ with non-null probabilities. In practice we are dealing
with observed system configurations with non-null probabilities, and
will act as if $X'=X$.%
}:

\begin{equation}
\mathbb{E}[U]=\sum_{x\in X}p\left(x\right)\sum_{z\in Z}U\left(y_{x},z\right)p\left(\left.z\right|x\right)\label{eq:Expected_Utility}
\end{equation}

The goal is usually to find a function $y_{x}$ that maximises the
expected utility: this would correspond to making the best predictions
on average (and implicitly acting accordingly). By analogy with causal
states \cite{CrutchfieldStatComp} and ε-machines, we now cluster
together configurations $x$ according to their statistical properties
and look at conditions for which these clustering lead to maximal
expected utility $\mathbb{E}[U]$.

\subsection{\label{sub:Definitions}Equivalence relations}

$\mathbb{E}[U]$ is maximal when each term $T\left(x,y\right)=p\left(x\right)\sum_{z\in Z}U\left(y,z\right)p\left(z|x\right)$
is maximal (see Eq. \ref{eq:Expected_Utility}). Since $p(x)$ is
constant for a given $T\left(x,y\right)$, and assuming we can choose
the $y$ for each $x$ independently, maximising $T$ is equivalent
to maximising each $\sum_{z\in Z}U\left(y,z\right)p\left(z|x\right)$.
Let us note $\mathbb{U}(y|x)=\mathbb{E}_{z\in Z}[U(y,z)|x]=\sum_{z\in Z}U\left(y,z\right)p\left(z|x\right)$,
the expected utility of choosing the prediction $y$ for a given $x$. 

Another assumption is implicit in this argumentation: that making
a decision does not modify the system. The weather forecasting example
in the introduction falls in this category. However, sometimes taking
a decision modifies the system. For example, when monitoring a patient's
health in order to decide whether to administrate a drug or not. In
that case we have to rely on approximations (usually an additional
assumption that the change is effective only at a different time scale
than that of the observations) so we can still aggregate them on a
recent past sliding window. Alternatively, other frameworks like Interactive
Learning \cite{infoInterLearn} might be better suited for these situations.

Let us now recall the causal states construction \cite{ShaliziThesis}:
\begin{description}
\item [{Causal~state~equivalence~relation:}] $x_{1}\overset{c}{\equiv}x_{2}$
if, and only if, the conditional distributions $P(Z|x_{1})=P(Z|x_{2})$
are the same. The equivalence classes $\sigma(x)=\left\{ w:P(Z|w)=P(Z|x)\right\} $
are called the \emph{causal states}. See Section \ref{sub:Causality_and_predictions}
for a discussion on this term.
\end{description}
By analogy with the causal states construction, let us now define
the following equivalence relations:
\begin{description}
\item [{Utility~equivalence~relation:}] $x_{1}\overset{u}{\equiv}x_{2}$
if, and only if, $\max_{y\in Z}\mathbb{U}\left(y|x_{1}\right)=\max_{y\in Z}\mathbb{U}\left(y|x_{2}\right)$.
That is, the maximal expected utility is the same at points $x_{1}$
and $x_{2}$, even if the sets of optimal predictions $Y\left(x_{1}\right)=\mathrm{argmax}_{y\in Z}\mathbb{U}\left(y|x_{1}\right)$
and $Y\left(x_{2}\right)$ that induce this utility might differ for
$x_{1}$ and $x_{2}$.
\item [{Prediction~equivalence~relation:}] $x_{1}\overset{p}{\equiv}x_{2}$
if, and only if, $\mathrm{argmax}_{y\in Z}\mathbb{U}\left(y|x_{1}\right)=\mathrm{argmax}_{y\in Z}\mathbb{U}\left(y|x_{2}\right)$.
That is, the sets of optimal predictions $Y\left(x_{1}\right)=Y\left(x_{2}\right)$
are the same, even if the utility induced by these predictions might
differ for $x_{1}$ and $x_{2}$.
\end{description}
Let us call \emph{iso-utility states} $\lyxmathsym{υ}\in\lyxmathsym{Υ}$
and \emph{iso-prediction states} $\lyxmathsym{ψ}\in\lyxmathsym{Ψ}$,
the partitions of $X$ corresponding to these equivalence relations:
$\lyxmathsym{υ}(x)=\left\{ x':x'\overset{u}{\equiv}x\right\} $ and
$\lyxmathsym{ψ}(x)=\left\{ x':x'\overset{p}{\equiv}x\right\} $.

Let us call \emph{decisional states} $\lyxmathsym{ω}\in\lyxmathsym{Ω}$
the intersection of both: $\lyxmathsym{ω}(x)=\left\{ x':x'\overset{p}{\equiv}x\textrm{ and }x'\overset{u}{\equiv}x\right\} $.
When both the expected utility and the optimal predictions are the
same, we assume the decisions that are taken on the system are the
same, hence the name. In other words, we suppose the utility function
encodes all that a user needs to take a decision.

These equivalence relations partition the configuration space $X$
into clusters, with the corresponding properties common to all points
in the cluster. It should be noted that $\mathbb{E}[U]$ as defined
on the whole space does not consider which specific decisional state
the process is in. Knowing which is the current cluster for any given
point $x$ allows us to refine the expected utility to a local $\mathbb{E}_{\lyxmathsym{υ}}[U]$
(with $x\in\lyxmathsym{υ}$ the iso-utility state) and which decision
to take to reach this utility (by refining again to the decisional
state). Section \ref{sub:Transition-graphs} details how to derive
notions of complexity from these local expected values.

\subsection{Relation between the causal, iso-utility, iso-prediction and decisional
states}

\begin{figure}
\begin{centering}
\includegraphics[width=1\columnwidth]{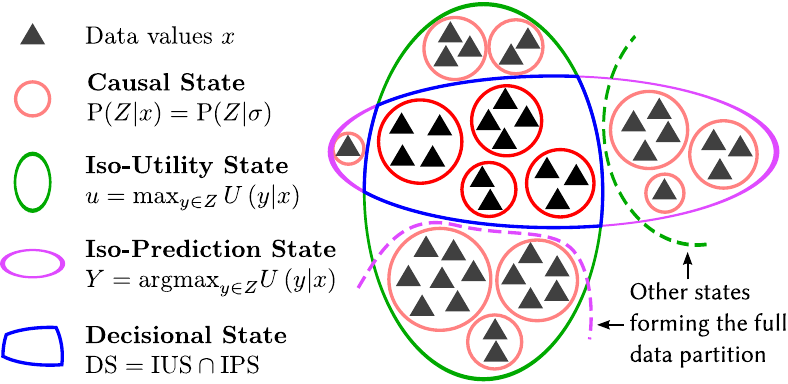}
\par\end{centering}

\caption{Relations between the different states. \label{fig:StatesRelation}}
\end{figure}

Let $x_{1}$ and $x_{2}$ be in the same causal state \foreignlanguage{english}{$\sigma$}.
Then by definition of the causal states $P(Z|x_{1})=P(Z|x_{2})$.
In that case, the expected utility of any prediction $y\in Z$ is
the same for $x_{1}$ and $x_{2}$: $\mathbb{U}(y|x_{1})=\sum_{z\in Z}U\left(y,z\right)p\left(\left.z\right|x_{1}\right)=\sum_{z\in Z}U\left(y,z\right)p\left(\left.z\right|x_{2}\right)=\mathbb{U}(y|x_{2})$.
Therefore the optimal predictions and induced utilities are the same:
$x_{1}\overset{p}{\equiv}x_{2}$ and $x_{1}\overset{u}{\equiv}x_{2}$,
and so is the combination of both.

Thus the causal states sub-partition both the iso-utility, iso-prediction
and decisional states.

The converse is not true: we can have two distinct causal states $\sigma_{1}$
and $\sigma_{2}$ with the same maximum value of $\mathbb{U}(y|x)=\sum_{z\in Z}U\left(y,z\right)p\left(\left.z\right|x\right)$
at the same $y$ points, but with different $p\left(\left.z\right|x\right)$
for at least one $z\in Z$.

Figure \ref{fig:StatesRelation} shows the relations between the different
states defined on the process.

\subsection{Transition graphs\label{sub:Transition-graphs}}

\begin{figure}
\begin{centering}
\includegraphics[width=1\columnwidth]{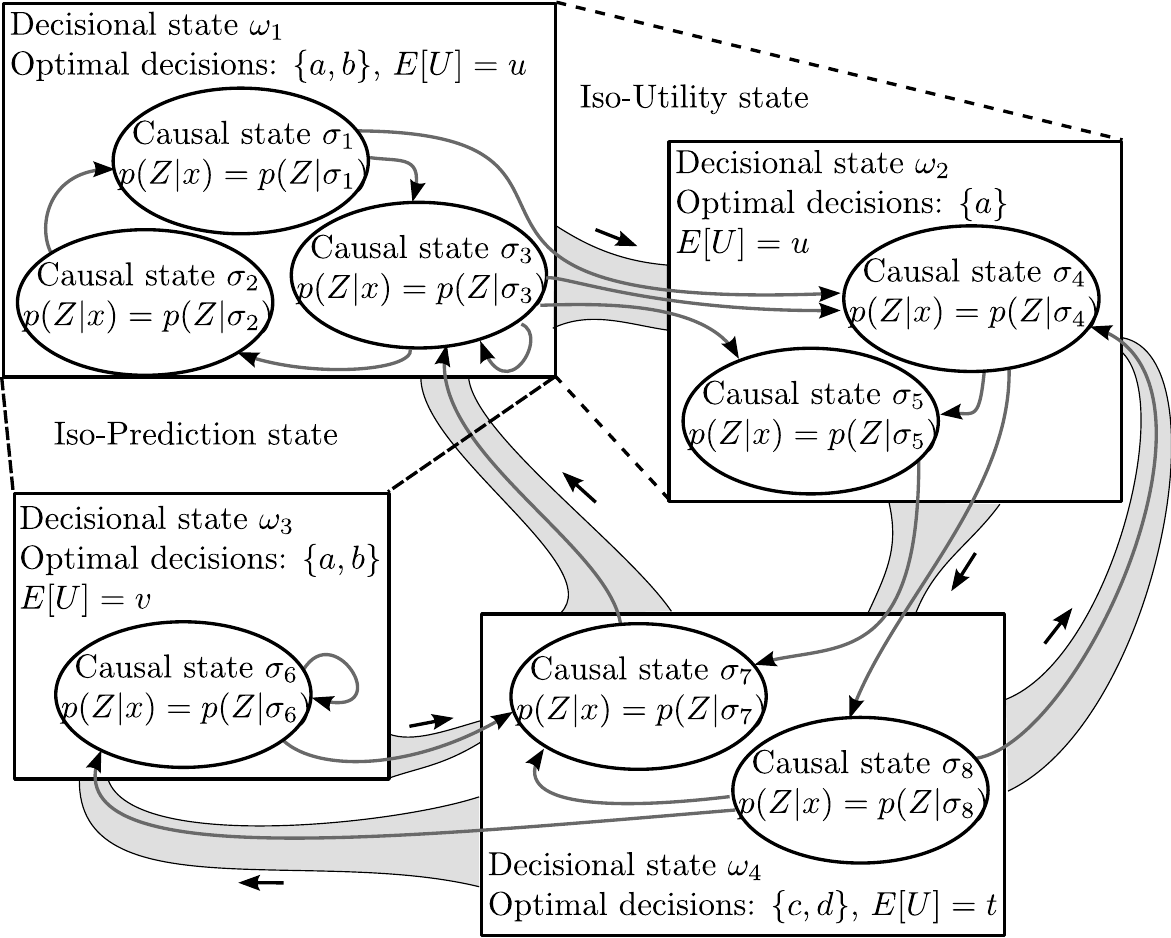}
\par\end{centering}

\caption{Decisional states transition graph on top of the ε-machine.\label{fig:Decisional-graph}}
\end{figure}

In the discrete case the causal states form a deterministic automaton,
the ε-machine \cite{CrutchfieldStatComp}. Since the iso-utility,
iso-prediction, and decisional states are coarser partitions than
the causal states each of them comprises one or more nodes of the
ε-machine automaton. The transitions between these nodes are either
internal to the coarser state, in which case they are ignored (or
represented as self-loops), or lead to another coarser state. 

But the ε-machine is a deterministic automaton: for each initial causal
state $\sigma_{1}$ and each discrete symbol $a\in\mathcal{A}$, in
a discrete scenario with alphabet $\mathcal{A}$, there is at most
one transition from $\sigma_{1}$ labeled by $a$ and it leads to
a unique causal state $\sigma_{2}$, possibly the same as $\sigma_{1}$.
If $\sigma_{1}$ and $\sigma_{2}$ belong to different coarser states,
the transition is also present at the coarser level. However there
may be several transitions with different labels between the same
coarser states: in Fig. \ref{fig:Decisional-graph} this is the case
for the transitions $\sigma_{4}\rightarrow\sigma_{8}$ and $\sigma_{5}\rightarrow\sigma_{7}$.
These transitions are not distinguishable at the coarser level: both
are included in $\omega_{2}\rightarrow\omega_{4}$. We therefore loose
the symbol labeling of the transitions at the coarser level. Figure
\ref{fig:Decisional-graph} shows decisional states as a coarser level
of causal states of an underlying ε-machine, with iso-utility and
iso-prediction states on top of the decisional states.

Let $x$ be the current system configuration, and $\lyxmathsym{ω}$
the current system decisional state (so $x\in\lyxmathsym{ω}$). Assume
symbol $a\in\mathcal{A}$ is observed at this point. Then $xa=s\in X$
is the system configuration after the observation. Let $\lyxmathsym{γ}$
be the decisional state for $s$. Then $p(e_{\lyxmathsym{ω}\rightarrow\lyxmathsym{γ}})=\sum_{x\in\lyxmathsym{ω}}\sum_{a\in A}p\left(xa\in\lyxmathsym{γ}|x\in\lyxmathsym{ω}\right)$
is the probability of the transition event $e_{\lyxmathsym{ω}\rightarrow\lyxmathsym{γ}}$
from state $\lyxmathsym{ω}$ to state $\lyxmathsym{γ}$. The same
construction also works for iso-utility and iso-prediction states:
\begin{itemize}
\item Iso-utility state transitions are events that change the expected
utility, irrespectively of the implied symbols. Several causal states
might belong to the same iso-utility state, as depicted in Fig. \ref{fig:StatesRelation}.
\item Iso-prediction state transitions are events that change the possible
optimal prediction choice, with the same comment.
\item Decisional state transitions change at least one of the above.
\end{itemize}
An important consequence of this definition is that we loose the Markov
property of the coarser graph%
\footnote{Thanks to one of the anonymous reviewers for pointing this example
and highlighting this aspect of the coarser graphs.%
}. For example, in the Fig \ref{fig:Decisional-graph}, $p(\Omega_{t}=\omega_{1}|\Omega_{t-1}=\omega_{4})\neq p(\Omega_{t}=\omega_{1}|\Omega_{t-1}=\omega_{4},\Omega_{t-2}=\omega_{3})$
as in this later case (coming from $\omega_{3}$) the internal causal
state of $\Omega_{t-1}=\omega_{4}$ is necessarily $\sigma_{7}$.
Nevertheless, the coarser states are still defined as the minimal
supersets of causal states leading to the same decisions. Knowing
the epsilon machine in addition to the superstructure may lead to
improved predictions of the future transitions, but it does not change
the utility and optimal values associated to the current state.

In computational mechanics \cite{CrutchfieldStatComp} the mutual
information $C=I(x;\sigma)$ between a configuration $x$ and the
causal state $\sigma$ for $x$ is referred to as the statistical
complexity. In the discrete case, by definition $C=I(x;\sigma)=H(\sigma)-H(\sigma|x)$
with $H$ proper entropies (differential entropies in the continuous
case). But then $H(\sigma|x)=0$ by construction of the $\sigma$,
which in the discrete case leads to well-defined transition graphs.
Thus $C=H(\sigma)$ in this case, the amount of information necessary
to encode the epsilon machine. 

In the present context we might define by analogy a \emph{decisional
complexity} $D$ as the amount of information necessary to encode
the coarser level graph, \emph{given a utility function}. Once that
utility function is fixed we can compute the decisional states and
define $D=I(x;\lyxmathsym{ω})$ where $x$ is a configuration of
the system and $\lyxmathsym{ω}$ the decisional state for $x$.

Since $-p(\lyxmathsym{ω})\log_{2}p(\lyxmathsym{ω})=-\left(\sum_{\sigma\in\lyxmathsym{ω}}p(\sigma)\right)\log_{2}\left(\sum_{\sigma\in\lyxmathsym{ω}}p(\sigma)\right)\leq-\sum_{\sigma\in\lyxmathsym{ω}}p(\sigma)\log_{2}p(\sigma)$
because $-\log_{2}$ is monotonically decreasing, then $D\leq C$
in the discrete case.

The same construction also works for defining similar quantities:
\begin{itemize}
\item $P=I(x;\lyxmathsym{ψ})$ is the tentatively called here the optimal
prediction complexity.
\item $V=I(x;\lyxmathsym{υ})$ is the complexity of estimating the expected
utility from $x$.
\end{itemize}
Similarly to \cite{filters_ca_stat_comp} it is possible to define
a local statistical complexity measure corresponding to each state:
$C_{\sigma}=-log_{2}(p(\sigma))$ in bits, and correspondingly for
the other coarser states : $D_{\omega}=-log_{2}(p(\omega))$, similarly
for $P_{\lyxmathsym{ψ}}$ and $V_{\nu}$. With this definition the
global values are simply the expectation of the local values over
all states (ex: $D=\sum_{\omega}p(\omega)D_{\omega}$). The local
complexity measures are used for the examples in Section \ref{sec:Application-examples}.

\subsection{\label{sub:Interpretation-and-notes}Interpretation and notes}

Decisional states are equivalent to merging those causal states which
lead to the same decisions relatively to our utility function. In
this case the causal states have lost their maximality property due
to the fact we're only interested in making a prediction and not in
keeping the full conditional distributions. We have, in the general
case, clustered together the causal states that lead to the same optimal
predictions and maximal expected utility value, based on a given utility
function.

Conversely this defines an equivalence relation amongst utility functions:
Two utility functions $U_{1}$ and $U_{2}$ are equivalent when they
induce the same clustering of causal states into the decisional ones,
with the same expected values and optimal predictions. These utility
functions would induce the same decisions in a system: they are functionally
equivalent. Isomorphisms between utility functions leading to the
same predictions but with different utility values could also be defined:
these are transformations of utility functions that preserve the iso-prediction
states. Similarly, transformations could be defined that only preserve
the iso-utility states.

The transitions between the iso-utility states correspond to events
that provoke a change in the expected utility of the system. Identifying
these events might become a crucial practical application, for example
for detecting when the expected utility reaches a predefined threshold.

The transitions between the iso-prediction states correspond to events
that provoke a change in the optimal predictions that can be chosen.
Similarly, a user might be interested in monitoring these changes,
for example, to maintain the current action as long as it is appropriate
(as long as it matches one of the possible predictions for the system's
evolving iso-prediction state).

The hypothesis made here is that when the cost/utility is defined
in terms of a functional (high-level) value, when it has a signification
in high-level terms, then the transition events also correspond to
interesting high-level objects to look at. Ignoring the internal transitions
and keeping only the coarser-level transitions also leads to a time-scale
change in practice. This might form the basis for an automated search
for meaningful events in a given system's evolution directly at a
higher-level than that of its constituents.

In any case, the utility function encodes external information not
available in the original data. So long as one stays with causal states,
only information present in the low-level data can be obtained. Much
like introducing a prior in a Bayesian framework, here the utility
function can be seen as encoding an a priori information not available
in the original data. This has at least three consequences from an
emergentist point of view:
\begin{itemize}
\item The causal states represent the finest scale at which we can meaningfully
associate a utility function and take decisions. Since the decisional
states are supersets of the causal states, then any partition of the
data defined with respect to a utility function cannot go below that
scale, whatever the chosen utility function.
\item Macro-level information needs not be computationally reducible \cite{synthesis_complexity}
to the lower level in order to be incorporated: The utility function
is defined on $Z^{2}$, not $X$, and it can possibly be incompressible,
stated as a value table and not explicitly computed in terms of the
lower-level scale. The data $x\in X$ is then clustered into sets
which need not have a meaning defined at that level.
\item If the hypothesis that {}``emergent structures are sub-machines of
the ε-machine'' \cite[sec. 11.2.2]{ShaliziThesis} is correct, then
the decisional states are the emergent structures corresponding to
a given utility function. Rather than looking for emergent entities
directly we might then encode our knowledge in a utility function,
and look at the decisional states in order to find good emergent entity
candidates. If these do not suffice, we might then refine the utility
function iteratively.
\end{itemize}
Finally, it should be noted the utility function is not the only source
of external introduction of knowledge in the system. Additional assumptions
are made either implicitly or explicitly if the system is able to
generalise to unknown values. For example, the hypothesis that $P(Z|x)$
can be decomposed using kernels or Bayesian networks could be one
such assumption. The accuracy of the proposed method for finding decisional
states depends on how well these extra assumptions are verified, independently
of the chosen utility function.

The algorithm proposed in Section \ref{sec:Algo} does not handle
the verification of preconditions, which are expected to be performed
by the user depending on the context (ex: nature of the data). However
the reference implementation (link given in Appendix) is fully generic
and allows testing different sampling and generalisation methods if
needed.

\section{\label{sec:Examples}Analytic examples}

\subsection{Example 1: when bad predictions are useless}

In this subsection utility is given to a prediction only if it is
correct; otherwise, the prediction is declared useless: $U(z,z)=1$
and $U(y,z\neq y)=0.$ In a continuous scenario the delta function
$U(y,z)=\lyxmathsym{δ}(y,z)$ is used instead.

From Section \ref{sub:Definitions}:

\[
\mathbb{U}\left(y|x\right)=\sum_{z\in Z}U\left(y,z\right)p\left(z|x\right)
\]

then becomes:

\[
\mathbb{U}\left(y|x\right)=p\left(\left.y\right|x\right)
\]
The set $Y(x)$ of predictions $y$ realising an optimal gain becomes:
\[
Y(x)=\left\{ y:\, p\left(\left.y\right|x\right)=max_{z\in Z}\, p\left(\left.z\right|x\right)\right\} 
\]

And Eq. \eqref{eq:Expected_Utility} leads to:

\[
\mathbb{E}_{max}[U]=\sum_{x\in X}p\left(x\right)\, max_{z\in Z}\, p\left(\left.z\right|x\right)
\]

But for each causal state $\sigma\subset\lyxmathsym{ω}$ in each
decisional state $\lyxmathsym{ω}$ the conditional probability $P\left(\left.Z\right|x\right)=P\left(\left.Z\right|\sigma\right)$
is the same for all $x\in\sigma$. The decisional states are found
by gathering causal states with the same maxima points $y$ for $P\left(\left.Z\right|\lyxmathsym{ω}\right)$.
We can then write in this special case the above formula as:

\[
\mathbb{E}_{max}[U]=\sum_{\lyxmathsym{ω}}\sum_{\sigma\subset\lyxmathsym{ω}}p(\sigma)p\left(\left.y_{\lyxmathsym{ω}}\right|\sigma\right)
\]
 where $y_{\lyxmathsym{ω}}$ is taken as any maxima of $p(Z|\sigma)$
common to all $\sigma\subset\lyxmathsym{ω}$.

Under the condition $U(z,z)=1$ and $U(y,z\neq y)=0$ (or $U(y,z)=\lyxmathsym{δ}(y,z)$
in the continuous case) the full conditional probability distributions
$p(Z|x)$ do not matter, what's important is that these distributions
peak at the same maxima.

\subsection{\label{Loss-err-squared}Example 2: Loss defined by error squared}

This section investigates the case where the loss function $L\left(y,z\right)$
can be written as a squared difference between the actual event and
the prediction: $L\left(y,z\right)=\left(z-y\right)^{2}$, provided
this operation is meaningful in $Z$.

We re-develop the treatment from \cite[section 1.5.5]{BishopML} in
our new context:

With the above loss function Eq. \ref{eq:Expected_Utility} becomes:

\[
\mathbb{E}[L]=\int_{z\in Z}\int_{x\in X}L\left(y_{x},z\right)p(x,z)\mathrm{d}x\mathrm{d}z
\]

\[
\mathbb{E}[L]=\int_{z\in Z}\int_{x\in X}\left(z-y_{x}\right)^{2}p(x,z)\mathrm{d}x\mathrm{d}z
\]

As in Section \ref{sub:General-case}, the goal is to find an $y_{x}$
function that minimises the expected loss: $\mathbb{E}_{min}[L]$.

The extrema of $\mathbb{E}[L]$ are given by the functional equation
$\frac{\partial\mathbb{E}[L]}{\partial y_{x}}=0$, with:

\[
\frac{\partial\mathbb{E}[L]}{\partial y_{x}}=2\int_{z\in Z}\left(z-y_{x}\right)p(x,z)\mathrm{d}z
\]

Solving $\frac{\partial\mathbb{E}[L]}{\partial y_{x}}=0$ gives:

\[
\int_{z\in Z}z\, p(x,z)\mathrm{d}z=y_{x}\int_{z\in Z}p(x,z)\mathrm{d}z
\]

So except for a set of $x$ with null probability mass:

\[
p(x)\int_{z\in Z}z\, p(z|x)\mathrm{d}z=y_{x}p(x)
\]

\begin{equation}
y_{x}=\mathbb{E}_{z\in Z}[z|x]\label{eq:leastSquaresSol}
\end{equation}

For a given causal state $\sigma$, $P(Z|x)$ is the same for all
$x\in\sigma$ so we can write $y_{\sigma}=\mathbb{E}_{z\in Z}[z|\sigma]$.

The decisional states are in this example obtained by clustering together
the causal states with the same expected value of $z$ within the
state.

These results are obtained because the utility function can be treated
analytically; in the general case we do not have such simple formula
available. The next section presents an algorithm that can infer the
structure of the decisional states from observed data and numerical
integration.

\section{\label{sec:Algo}Estimating the decisional states from data}

\subsection{\label{sub:AlgoSetup}General presentation of the algorithm}

There are two main critical tasks the algorithm must perform:
\begin{itemize}
\item Estimating the probability distributions $P(Z|x)$ from data. The
probability distribution estimator is responsible for providing values
for unobserved data (generalisation ability). It might use all available
observations: $\hat{P}(Z|x)=F(\mathcal{O})$, where $\mathcal{O}=\left\{ \left(\boldsymbol{x_{i}},\boldsymbol{z_{i}}\right)_{i=1\lyxmathsym{…}N}\right\} $
represents the data in the form of observation pairs $\left(\boldsymbol{x_{i}},\boldsymbol{z_{i}}\right)$,
and $F$ a generic function.
\item Clustering $X$ into causal, iso-prediction, iso-utility and decisional
states according to the user needs. This implies as a sub-task estimating
the maxima for $\mathbb{U}\left(y|x\right)$. The first step is to
build $\hat{\mathbb{U}}(y|x)=\int_{Z}\hat{p}(z|x)U(y,z)\mathrm{d}z$
with a user-provided integrator and the utility function. Then, an
optimiser might be invoked so as to compute $Y(x)=\mathrm{Argmax}_{y\in Z}\hat{\mathbb{U}}(y|x)$.
\end{itemize}
So, in summary, the user must provide (explicitly or implicitly using
the reference implementation default choices):
\begin{itemize}
\item A probability density estimator $\hat{P}(Z|X)$ from data observations
$\mathcal{O}$.
\item A utility function $U$ acting on $Z^{2}$ with $Z$ the space of
predictions.
\item An integrator for computing the expected value of $U$ with respect
to the estimated density.
\item A multi-modal optimiser in order to compute $Y(x)=\mathrm{Argmax}_{y\in Z}\hat{\mathbb{U}}(y|x)$.
\item A clustering algorithm for gathering probability distributions (for
causal states), utility values (for iso-utility states), or similar
sets $Y(x)$ (for iso-prediction states). Decisional states are found
by intersection of the iso-utility and iso-prediction states.
\item Optionally, the user may associate a symbol to each transition $\left(\boldsymbol{x_{t}},\boldsymbol{x_{t+1}}\right)$.
This step is detailed in Section \ref{sub:Determinism-vs-CSSR}, once
the main algorithm is explained and we can see why and when this step
may become necessary.
\end{itemize}
Figure \ref{fig:Dstates-rec-algo.} recapitulates these points and
shows the algorithm steps that will be detailed in the next sections.
Readers not used to generic programming might be suprised by the functional
inputs: The reference implementation proposes default choices for
these functions, but the user is free to provide any equivalent replacement
code if so desired. The algorithm in Fig. \ref{fig:Dstates-rec-algo.}
is described in general terms but it is well-defined and a concrete
reference implementation in C++ is provided, see the Appendix.

\begin{figure}
\noindent {\footnotesize Data inputs:}{\footnotesize \par}

\noindent {\footnotesize – Pairs of observations $\mathcal{O}=\left\{ \left(\boldsymbol{x_{i}},\boldsymbol{z_{i}}\right)_{i=1\lyxmathsym{…}N}\right\} $;}{\footnotesize \par}

\noindent {\footnotesize – (Optional): Symbols $\left(\boldsymbol{s_{i}}\right)_{i=1\lyxmathsym{…}N-1}$
for each transition $\left(\boldsymbol{x_{i}},\boldsymbol{x_{i+1}}\right)_{i=1\lyxmathsym{…}N-1}$;}{\footnotesize \par}

\noindent {\footnotesize – Parameters for the functional inputs (ex:
threshold for matching probability distributions).}{\footnotesize \par}

\bigskip{}

\noindent {\footnotesize Functional Inputs:}{\footnotesize \par}

\noindent {\footnotesize – A probability density estimator $PDE$
such that $\hat{P}(Z|X)=PDE(\mathbb{\mathcal{O}})$. The distribution
type is user-defined;}{\footnotesize \par}

\noindent {\footnotesize – A utility function $U:\, Z^{2}\mapsto\mathbb{R}$;}{\footnotesize \par}

\noindent {\footnotesize – An integrator $Integ$ over $Z$;}{\footnotesize \par}

\noindent {\footnotesize – A multi-modal optimiser $Argmax$ over
$Z$;}{\footnotesize \par}

\noindent {\footnotesize – A clustering algorithm $C1$ acting on
probability densities $\hat{P}(Z|X)$;}{\footnotesize \par}

\noindent {\footnotesize – A clustering algorithm $C2$ over subsets
of $Z$;}{\footnotesize \par}

\noindent {\footnotesize – A clustering algorithm $C3$ over $\mathbb{R}$.}{\footnotesize \par}

\bigskip{}

\noindent {\footnotesize Algorithm:}{\footnotesize \par}
\begin{enumerate}
\item {\footnotesize Build the density estimates $\hat{P}(Z|\boldsymbol{x_{i}})$
for each $\boldsymbol{x_{i}}$ in the data set using $PDE$.}{\footnotesize \par}
\item {\footnotesize Cluster the density estimates using $C1$ into causal
states $\hat{\sigma}$.}{\footnotesize \par}
\item {\footnotesize (Optional) Refine the estimates $\hat{\sigma}$ and
loop to step 2 using the symbols $\left(\boldsymbol{s_{i}}\right)_{i=1\lyxmathsym{…}N-1}$.
See Section \ref{sub:Determinism-vs-CSSR}.}{\footnotesize \par}
\item {\footnotesize Average out $\hat{P}(Z|\hat{\sigma})=avg_{\boldsymbol{x_{i}}\in\sigma}\hat{P}(Z|\boldsymbol{x_{i}})$.
See Section \ref{sub:Clustering}.}{\footnotesize \par}
\item {\footnotesize Compute $Y(\hat{\sigma})=Argmax_{y}\, Integ_{z}\left(U(y,z)\hat{P}(z|\hat{\sigma})\right)$
for each causal state estimate $\hat{\sigma}$, retaining the utility
$U(\hat{\sigma})$ obtained for these maxima.}{\footnotesize \par}
\item {\footnotesize Cluster the causal states estimates using $Y(\hat{\sigma})$
and $C2$ into iso-prediction estimates $\hat{\lyxmathsym{ψ}}(\hat{\sigma})\in\hat{\lyxmathsym{Ψ}}$.}{\footnotesize \par}
\item {\footnotesize Cluster the causal states estimates using $U(\hat{\sigma})$
and $C3$ into iso-utility estimates $\hat{\lyxmathsym{υ}}(\hat{\sigma})\in\hat{\lyxmathsym{Υ}}$.}{\footnotesize \par}
\item {\footnotesize Intersect $\hat{\lyxmathsym{Ψ}}\cap\hat{\lyxmathsym{Υ}}$
into decisional states $\hat{\lyxmathsym{Ω}}$ that partition $X$.}{\footnotesize \par}
\item {\footnotesize (Optional) Produce the transition graphs, and the ε-machine
if the symbols are available.}{\footnotesize \par}
\item {\footnotesize (Optional) Compute the global complexities of the system
$C$, $D$, $P$, and $V$ from section \ref{sub:Transition-graphs}.}{\footnotesize \par}
\item {\footnotesize (Optional) For each $\boldsymbol{x_{i}}$, compute
the local complexity equivalents of $C$, $D$, $P$, and $V$ at
this point (see Section \ref{sub:Transition-graphs}).}{\footnotesize \par}
\end{enumerate}
\caption{Decisional state reconstruction algorithm.\label{fig:Dstates-rec-algo.}}
\end{figure}

\subsection{\label{sub:Kernel-density-estimation}Kernel density estimation}

This section describes one way to perform Step 1 in Fig. \ref{fig:Dstates-rec-algo.}.

A discrete probability estimator is suitable for small $X$ spaces
where a sufficient amount of data was observed, so that $P(Z|X)$
can be reliably estimated by counting occurrences of all $x$ and
$z$. For larger spaces or when unknown or continuous $X$ might be
encountered, the system must be able to generalise. We now present
the case for a Kernel Density Estimation (KDE) \cite{KDE} of the
probability density $\hat{P}(Z|X)=F(\mathbb{\mathcal{O}})$.

In general, the kernel $K(a;b)$ with $a$ and $b$ in the joint space
$\{a,b\}\subset X\times Z$ is not separable: The density estimate
is $\hat{p}(x,z)\propto\sum_{i=1}^{N}K(\boldsymbol{x_{i}},\boldsymbol{z_{i}};x,z)$,
summing over all observation pairs $\left(\boldsymbol{x_{i}},\boldsymbol{z_{i}}\right)\in\mathbb{\mathcal{O}}$.
In the particular case of separable kernels for the configuration
space $X$ and the prediction space $Z$ we have instead: $\hat{p}(x,z)\propto\sum_{i=1}^{N}K^{x}(\boldsymbol{x_{i}};x)K^{z}(\boldsymbol{z_{i}};z)$.
Even when the kernel is separable the user may benefit from the joint
kernel approach: For analysing time series it is natural to consider
a moving window of $dim(X\times Z)$ values and perform the density
estimation on the joint space. In another example in Section \ref{sub:Image-segmentation}
an image is considered as the limit distribution of a Markov Random
Field \cite{AwateThesis}, and the density estimation is also performed
on the joint space (with $Z$ being in that example the space of pixel
values and $X$ the space of pixel neighbourhoods).

In any case, the conditional probability density is estimated by integrating
out the $\hat{p}(x)$ factor over $Z$: $\hat{p}(z|x)=\hat{p}(x,z)/\int_{\lyxmathsym{ζ}\in Z}\hat{p}(x,\lyxmathsym{ζ})$.
Several sampling mechanisms are provided over $Z$ for the integration,
including exhaustive listing of $Z$ for small search spaces. The
adequate method depends on the particular user application. 

Computing the causal states (and the other states built on top of
the causal states) only requires the conditional distributions, and
not the joint ones. Without loss of generality it is thus possible
to request that $K(a,a)=1$ with $a$ being $x$ or $z$ or the joint
data $(x,z)$ depending on the above cases. For example, the radial
basis function $K(a,b)=e^{-\|a-b\|^{2}/h}$, with $h$ the kernel
width. Indeed, dividing by $\hat{p}(x)$ absorbs the change of scale.
Better numerical accuracy is however achieved by requesting $K(a,a)=1$,
especially in high dimensions where the multivariate Normal kernel
would lead to very small $K(a,a)$.

The discrete case is recovered when choosing the delta function as
a kernel. In that case, similar observations $(\boldsymbol{x_{i}},\boldsymbol{z_{i}})$
are effectively summed up for a given $\boldsymbol{x_{i}}$ and the
probability estimator is an histogram. In practice it is preferable
to use a specialised discrete estimator implementation for efficiency
reasons.

Finally, the kernel width $h$ can be chosen according to a variety
of estimators from the data \cite{DuongThesis}. In practice it has
been observed that results ultimately depends on the final task for
which the algorithm is applied to. $h$ is then considered as a free
parameter, which can be determined for example by cross-validation
or by using a genetic algorithm. This gave the best results for classification
tasks based on the decisional complexity feature (ex: classification
of EEG time series \cite{eegdstates}). An hypothesis is that while
the kernel width $h$ found this way does not realise an a priori
form of optima (like the AMISE \cite{DuongThesis}), it realises an
a posteriori ideal compromise between bias and variance in the estimated
density for the particular task the algorithm is applied to. This
is similar to the approach in \cite{Seismic_CSSR_GA} except that
we have reduced the meta-parameter search to $h$ and got rid of the
histogram boundaries by using a KDE.

The default implementation proposes a reasonable choice based on the
average distance between nearest data points, from which the aforementioned
cross-validation and search techniques can build on.

\subsection{Using the probability estimates}

Two operations are performed using $\hat{P}(Z|x)$:
\begin{itemize}
\item Comparison: We need to check whether $\hat{P}(Z|x_{1})$ and $\hat{P}(Z|x_{2})$
are similar for clustering or not $x_{1}$ and $x_{2}$ in the same
causal state.
\item Expectation: We need to estimate the expected utility of a prediction
$y\in Z$ for a given $x\in X$: $\hat{\mathbb{U}}(y|x)=\int_{z\in Z}\hat{p}(z|x)U(y,z)$.
\end{itemize}
Comparison is handled by choosing a similarity measure between probability
distributions. The reference implementation proposes the $\lyxmathsym{χ}^{2}$
statistic, the Bhattacharyya, Variational and Harmonic mean distances,
and the Jensen-Shannon divergence \cite{distribution_distances}.
The Bhattacharyya distance is the default for the Kernel Density Estimation,
and a $\lyxmathsym{χ}^{2}$ test is the default for the discrete
case.

Let $S\subseteq Z$ be a set of sample points used over $Z$ for comparing
the probability distributions (possibly with $S=Z$ for an exhaustive
approach). Expectation of the utility for a candidate $y\in Z$ is
simply performed numerically over $Z$ at the chosen sample points
$S\subseteq Z$: $\hat{\mathbb{U}}(y|x)=\left(\sum_{s\in S}\hat{p}(s|x)U(y,s)\right)/\left(\sum_{s\in S}\hat{p}(s|x)\right)$.

\subsection{\label{sub:Clustering}Clustering}

Clustering of the causal states is performed directly by matching
the probability distributions as defined in the previous subsection.

For iso-utility and iso-prediction states an additional step is necessary:
Optimising $\hat{\mathbb{U}}(y|x)$ in order to find $Y(x)=\mathrm{Argmax}_{y\in Z}\hat{\mathbb{U}}(y|x)$.
Any multi-modal optimisation scheme can be invoked at this point.
Equivalent best predictions $y\in Y(x)$ must be found, so uni-modal
search schemes returning only one candidate are not adapted. Once
the prediction sets $Y(x)$ and the optimal utility values are computed
it is possible to cluster them.

False positives are when $x_{1}$ and $x_{2}$ are clustered together
when they are mathematically not equivalent, false negatives are when
the points are in different states when they should not. These risks
are minimised by providing more sample points to look at and by increasing
the data size. In the limit of an infinite number of data points and
samples, consistency is determined by the chosen approximate matchers
(ex: the Bhattacharyya distance in the previous section) and by whether
the data respects or not the mathematical assumptions needed for the
theory to work (ex: conditional stationarity of the $P(Z|x)$).

Additionally we can exploit the fact that causal states sub-partition
the decisional ones. If we compute the iso-utility, iso-prediction
and decisional states first we might then restrict the search for
similar probability distributions $P(Z|x_{1})$ and $P(Z|x_{2})$
to points $\left\{ x_{1},x_{2}\right\} \subset\lyxmathsym{ω}$ within
each decisional state $\lyxmathsym{ω}$. 

If we compute the causal states first (as in Fig. \ref{fig:Dstates-rec-algo.}),
we might use representative $\hat{P}(Z|\sigma)$ distributions for
each causal state, by averaging all $\hat{P}(Z|x)$ for $x\in\sigma$
in order to reduce numerical discrepancies: $\hat{P}(Z|\sigma)=\mbox{avg}_{x\in\sigma}\hat{P}(Z|x)$.
The expected utility $\hat{\mathbb{U}}(y|x)$ is then set to $\hat{\mathbb{U}}(y|\sigma)$
for all $x\in\sigma$. This approach (causal states first) was found
to give better results in practice.

The reference implementation provides two clustering algorithms. The
first algorithm is simply the straightforward implementation of matching
each candidate entity to each currently found cluster, and maintaining
averages for the clusters as described above. Once each observation
pair was processed the clusters themselves are matched with each other.
This step reduces the risk of spurious clusters that appear as a side-effect
of the implicit ordering in which the data is processed. The data
may then be re-matched iteratively if so desired in $P$ passes. This
algorithm complexity is at worse $O(P(KN+K(K-1)/2)))$ with $K$ the
number of clusters and $N$ the data size. It is thus \emph{linear
in data size}, and faster than the second algorithm proposed below.
The inconvenient is a dependence on the data presentation order, although
this can be minimised by randomising the data presentation order and
by performing several passes as aforementioned. See also the discussion
in the reference \cite{filters_ca_stat_comp} where their implementation
in Object Caml uses a similar but simpler clustering technique for
computing causal states. The leading Causal State Splitting Reconstruction
(CSSR) algorithm \cite{CSSR} in the domain also uses data order randomisation
and argues for consistency in the limit of a large number of observations.
See also \cite{drrn2} where the author proposes an incremental version
of this first algorithm that is additionnally able to handle data
on sliding windows in order to cope with slowly non-conditional stationnarity
systems.

The second provided algorithm is a single-linked hierarchical clustering
with a complexity in $O(N(N-1)/2)$. This is equivalent to finding
connected components with respect to the given match predicate, similar
to the DBSCAN \cite{DBSCAN} algorithm except that we must label each
value to a cluster (DBSCAN leaves out some values as noise). Depending
on the application one may prefer this algorithm to the first despite
of its worse cost for the following reasons:
\begin{itemize}
\item Arbitrary cluster shapes, unlike the first algorithm where clusters
are balls around the average value according to the similarity measure.
\item The clusters are stable with respect to the data presentation order.
\item Occam's razor: we want to find the simplest model able to handle the
data. Connected components maximise the clusters size by gathering
data when a matching path is found between them. This leads to a minimal
number of states in the discrete case while ensuring that data in
different states do not match (consistency), hence minimal statistical
$C=H(\sigma)$ or decisional $D=H(\lyxmathsym{ω})$ complexity values.
The downside is a sensitivity to the single-link effect: a single
error in the data may join several clusters, hence increasing the
number of the aforementionned false positives.
\item An interpretation for the continuous case. Connected components ensure
$d(a,b)>\lyxmathsym{Δ}$ for $a$ and $b$ points in different clusters,
$d$ a dissimilarity measure (ex: in utility values for iso-utility
states, on probability distributions for causal states, etc), and
$\lyxmathsym{Δ}$ a threshold for the mismatch between clusters.
This is equivalent to single-linked hierarchical clustering where
we cut the hierarchy at level $\lyxmathsym{Δ}$. In the continuous
case the transition graph construction fails on a continuum of infinitely
many nearby states. In that case connected components with threshold
$\lyxmathsym{Δ}$ ensure that the system state changed at least by
that amount when transiting from one point to the next in a different
component. For example when monitoring a system expected utility value,
a decision might be taken only when a sudden change is detected, but
not for a gradual change of the same magnitude.
\end{itemize}
Clustering is a research domain in itself and an important aspect
of data mining. The reference implementation proposes the above two
choices as their trade-off covers a vast range of usual cases. The
user is welcome to plug in a custom algorithm: thanks to the generic
nature of the reference implementation using C++ templates the clustering
part is independent of the rest of the computations.

\subsection{\label{sub:Determinism-vs-CSSR}Ensuring the ε-machine determinism}

Causal State Splitting Reconstruction (CSSR) \cite{CSSR} is the reference
algorithm for reconstructing ε-machines on discrete strings of symbols.
It works by recursively splitting the current causal state estimates
as the string length is increased. The consistency on shorter string
lengths is maintained while the causal states are refined to take
in account more symbols. In the limit it provably converges to the
true causal states.

In the present case we do not act on strings of symbols but on $(x,z)$
mappings. Hence it is not possible to refine iteratively the current
causal state estimates by enlarging the dimensions of $X$ and $Z$.
Yet the {}``symbols'' of discrete data are implicitly present in
the $(\boldsymbol{x_{i}},\boldsymbol{x_{i+1}})$ transitions when
monitoring the system (the index $i$ corresponds for example to ordered
time steps, but spatial transitions are possible as well). It would
be possible to recover a symbolic representation of the data set from
all such transitions%
\footnote{For a discrete finite data set, assume $x\in X$ is coded on $N$
bits. The difference between \foreignlanguage{english}{$\boldsymbol{x_{i}}$}
and $\boldsymbol{x_{i+1}}$ is always representable as a symbol in
an alphabet with size at most $2^{N}$.%
}, and apply CSSR if so desired. Here we directly cluster the system
configurations $x\in X$, not necessarily represented as strings of
symbols. For example, each $x\in X$ might correspond to a past light
cone (see Appendix A).

The drawback is that the proposed algorithm does \emph{not} so far
ensure that the resulting automaton is deterministic in terms of symbol
transitions, which is a condition for being a valid ε-machine \cite{ShaliziThesis,genNatPred}.
The labeled transitions between states can be recovered by looking
at the symbol suffix implied by passing from $\boldsymbol{x_{i}}$
to $\boldsymbol{x_{i+1}}$. But there is no guarantee at this point
that a given (state+symbol) combination always lead to the same state
deterministically.

Example: Suppose that $\boldsymbol{x_{i}}=aaba$ and $\boldsymbol{s_{i}}=abba$
are in the same causal state: $P(Z|\boldsymbol{x_{i}})$ and $P(Z|\boldsymbol{s_{i}})$
match and were clustered together with string length limited to depth
4. We observe that $\boldsymbol{x_{i+1}}=abac$ and $\boldsymbol{s_{i+1}}=bbac$,
with the same suffix $c$, and that $P(Z|\boldsymbol{x_{i+1}})$ and
$P(Z|\boldsymbol{s_{i+1}})$ do not match anymore and therefore were
not clustered in the same state. This is a violation of the ε-machine
determinism: from the same state and with the same symbol, the transition
leads to different states. Yet this case is possible when clustering
independently $\boldsymbol{x_{i}},\boldsymbol{s_{i}},\boldsymbol{x_{i+1}},\boldsymbol{s_{i+1}}$
into their own states as we do.

For iso-utility, iso-prediction and decisional states this is not
a problem: As explained in Section \ref{sub:Transition-graphs} transitions
are determined in terms of changes in utility related quantities,
the string symbols are irrelevant in that case (in other words, the
distinct causal states of the underlying ε-machine would be merged
into the coarser level). For an ε-machine reconstruction however the
proposed algorithm needs to be augmented with an additional step.

The user can optionally express symbol values together with each $\boldsymbol{x_{i}}\rightarrow\boldsymbol{x_{i+1}}$
transition. These are used as constraints for the clustering algorithm
when they are available. The following procedure is implemented:
\begin{itemize}
\item Clustering is performed as described in Section \ref{sub:Clustering}.
\item After clustering, iterate the following steps:

\begin{itemize}
\item Split step: It might be that data values $\boldsymbol{x^{1}}$ and
$\boldsymbol{x^{2}}$ for the transitions $\boldsymbol{x^{1}}\overset{a}{\rightarrow}\boldsymbol{s^{1}}$
and $\boldsymbol{x^{2}}\overset{a}{\rightarrow}\boldsymbol{s^{2}}$
were clustered together, while $\boldsymbol{s^{1}}$ and $\boldsymbol{s^{2}}$
are not clustered together (see the above example). In that case,
the state containing $\boldsymbol{x^{1}}$ and $\boldsymbol{x^{2}}$
is split in order to restore determinism.
\item Merge step: If several $x_{i}=\boldsymbol{x}\overset{a}{\rightarrow}x_{i+1}=\boldsymbol{s}$
transitions are observed, with the same configuration value $\boldsymbol{x}\in X$
and symbol $a\in A$, then all corresponding $\boldsymbol{s}\in X$
are pre-clustered in the same state. Note that splitting acts on the
states of mismatching configurations before a transition, while merging
acts on the states of mismatches after a transition, so both can be
applied without undoing each other.
\item Break the loop in the case of incompatible constraints and there is
no convergence.
\end{itemize}
\end{itemize}
Convergence of the loop would effectively ensure determinism of the
reconstructed automaton in the perfect case where all distribution
estimations are exact.

Unfortunately this is not the case in practice. Indeed, clustering
from finite data is necessarily imperfect. If $x\in\sigma_{2}$ is
wrongly affected to causal state $\sigma_{1}$ then forcing symbol
determinism might create spurious states: $\sigma_{2}$ is erroneously
split until the transitions are consistent, while the source of the
inconsistency is not detected. Or similarly states are merged when
they should not.

We had to accept a threshold for clustering distributions together
(ex: a significance level for the Chi-Square test), due to the imperfect
distribution estimation. In turn, we have no choice but to accept
that some $x\in\sigma$ might be misclassified and might generate
spurious transitions. The same way we ignore small discrepancies in
distribution clustering, the solution is to ignore small discrepancies
in the automaton determinism. Formally:

Let $\sigma$ be a causal state, $a\in A$ a symbol in the alphabet
$A$. The automaton is deterministic when each time a data value $x\in\sigma$
is followed by the symbol $a$ then $s=xa$ falls in a unique causal
state $\lyxmathsym{φ}$, $\forall x\in\sigma$. When the automaton
is not deterministic there is instead a distribution $p(\lyxmathsym{Φ}|\sigma,a)$
with $\lyxmathsym{φ}\in\lyxmathsym{Φ}$.

We propose here to set a threshold $\lyxmathsym{θ}=1-\lyxmathsym{ε}>\frac{1}{2}$
for ignoring small discrepancies up to $\lyxmathsym{ε}$: When $\exists\lyxmathsym{φ}/p(\lyxmathsym{φ}|\sigma,a)>\lyxmathsym{θ}$
then the unique such $\lyxmathsym{φ}$ is taken as the automaton
transition. This threshold is completely separate from the probability
of the transition itself.

Concretely, the split and merge step described above are applied only
on such transitions $\lyxmathsym{φ}$, the spurious transitions are
ignored.

\subsection{\label{sub:AlgoComplexity}Complexity of the algorithm}

Depending on the user context, one or the other of these tasks might
become the dominant algorithm cost:
\begin{itemize}
\item Estimating the probability distributions $\hat{P}(Z|x)$. For discrete
data the computation cost is simply $O(N)$, a matter of counting
the occurences of each $z\in Z$ for each $x$. But using the above
Kernel Density Estimation the complexity is roughly $O\left(N(M+q(N))\right)$
with $N$ the data size and $M$ the number of samples $s\in S\subset Z$
at which $\hat{p}(Z=s|x)$ is estimated. $q()$ is the cost of performing
a nearest neighbours query in the joint space (so negligible kernel
values are quickly eliminated, the worse-case limit of $q=O(N)$ is
the summing of all kernel values at all data points). The nearest
neighbors are estimated in a first pass in $O\left(Nq(N)\right)$
time and in the second pass the samples are used to build the distributions
in $O\left(NM\right)$ time. 
\item Clustering tasks. The two clustering algorithms described in Section
\ref{sub:Clustering} have a linear and a quadratic complexity with
respect to the dissimilarity measure.
\item Evaluating the utility function. For the analytical examples in the
Section \ref{sec:Examples}, $U(y,z)$ is simple enough so its evaluation
was not the main issue. However in a different scenario the algorithm
complexity might have to be defined in terms of the number of evaluations
of the cost function.
\item Optimising $\hat{\mathbb{U}}(y|x)$ in order to find $\hat{Y}(x)=\mathrm{Argmax}_{y\in Z}\hat{\mathbb{U}}(y|x)$.
An exhaustive search of $Z$ is only feasible for small discrete spaces.
Advanced multi-modal optimisation techniques might become necessary
and induce large computation times.
\end{itemize}
The memory requirements for running these computations might also
become a limiting factor. For example it might not be possible to
store all $\hat{P}(Z|x)$ distributions for each unique $x\in X$
present in the data set, especially if a large number of samples $s\in Z$
is needed (ex: the Monte-Carlo sampling error decreases as $O(1/\sqrt{|S|})$).

\section{\label{sec:Application-examples}Application examples}

The proposed algorithm is quite adaptable to various situations. Figure
\ref{fig:Dstates-rec-algo.} actually defines a familly of algorithms,
depending on how the functional inputs are implemented. For example,
estimating $\hat{P}(Z|x)$ is quite difficult in general, and depending
on the nature of the data (discrete or continuous) it can lead to
many algorithm variants. Similarly for the clustering task. As the
results are highly dependent on these crucial tasks, it is important
specify these functional inputs together with the algorithm in concrete
applications.

The following examples were thus choosen so as to highlight several
cases:
\begin{itemize}
\item The core part of the algorithm, reconstructing the ε-machine, with
comparison to a classical benchmark in the litterature \cite{CSSR}.
\item A toy application that is also compared to reference results in the
litterature \cite{filters_ca_stat_comp}. The goals are to highlight
how to use the light-cone formalism instead of symbolic series, as
well as to show to use of a utility fonction.
\item A larger application on a practical problem (edge detection) that
highlights how to apply the algorithm on a larger scale, as well as
the influence of the utility function.
\end{itemize}
In addition to these examples the algorithm was also applied to real
electroencephalogram (EEG) data in a separate paper \cite{eegdstates},
which is out of scope of the present introductionary material.

\subsection{\label{sub:evenProcess}Reconstruction of the Even process}

This first application example demonstrates the capability of performing
an ε-machine reconstruction. Since the ε-machine is the minimal and
optimal deterministic automaton for reproducing a process statistically,
and since the decisional states transition graph is a sub-machine
of the ε-machine (see Section \ref{sub:Interpretation-and-notes}),
the proposed algorithm needs to perform well on this task.

The Even process is used as a benchmark in \cite{CSSR}. A similar
experiment is conducted with the proposed algorithm for comparison.

\begin{figure}[h]
\begin{centering}
\includegraphics[width=0.8\columnwidth]{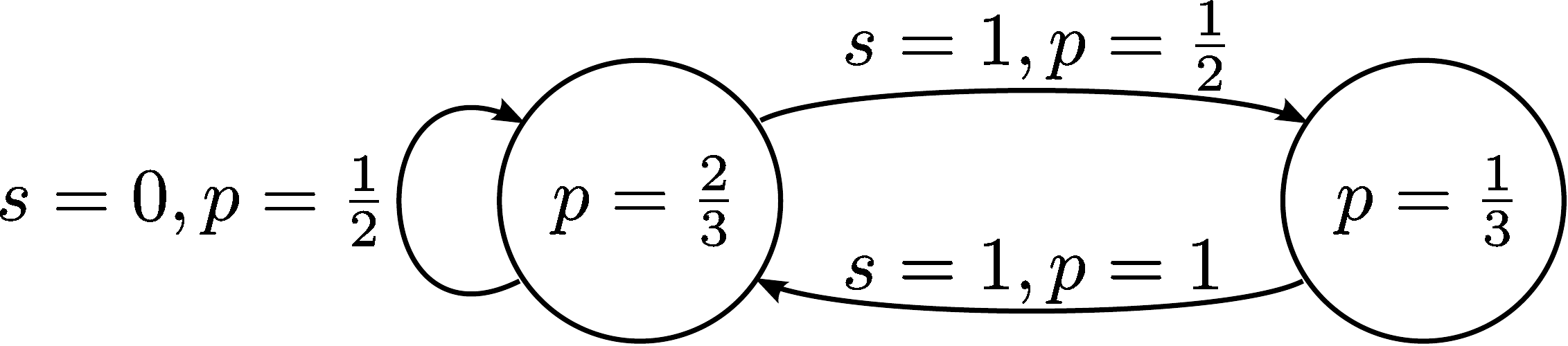}
\par\end{centering}

\caption{Definition of the Even process.\label{fig:Even-process-def}}
\end{figure}

The Even process consists of two states, and generates binary strings
where blocks of an even number of 1s are separated by an arbitrary
number of 0s. Despite its apparent simplicity the Even process does
not correspond to any finite state-output HMM \cite{DMarkovMachine},
and requires the power of an ε-machine to be reconstructed%
\footnote{Let's consider a naive state-output HMM emitting 0 in the left state
of Fig. \ref{fig:Even-process-def} and 1 in the right state. Let's
start with the left state. There may be an arbitrary number of 0 emitted.
When a transition is taken to the right state then a 1 is emitted
after the transition. The process then returns to the left state and
necessarily emits a 0. Adding a self-loop on the right state would
not satisfy the even number of 1 requirement. We could chain two states
emitting a 1 before returing to the left state, forcing the even number
two. But then we would need another chain for emitting four symbol
1 in a row, and so on. In fact, no finite state-output HMM can be
constructed for the Even process, while this is trivial with an edge-output
HMM: that is, an ε-machine.%
}. Figure \ref{fig:Even-process-def} shows the process states and
transitions.

Data is generated according to the Even process as a series of symbols.
The goal of the experiment is to reconstruct the underlying transition
graph from these observations.

\begin{figure}[h]
\begin{centering}
\includegraphics[width=0.8\columnwidth]{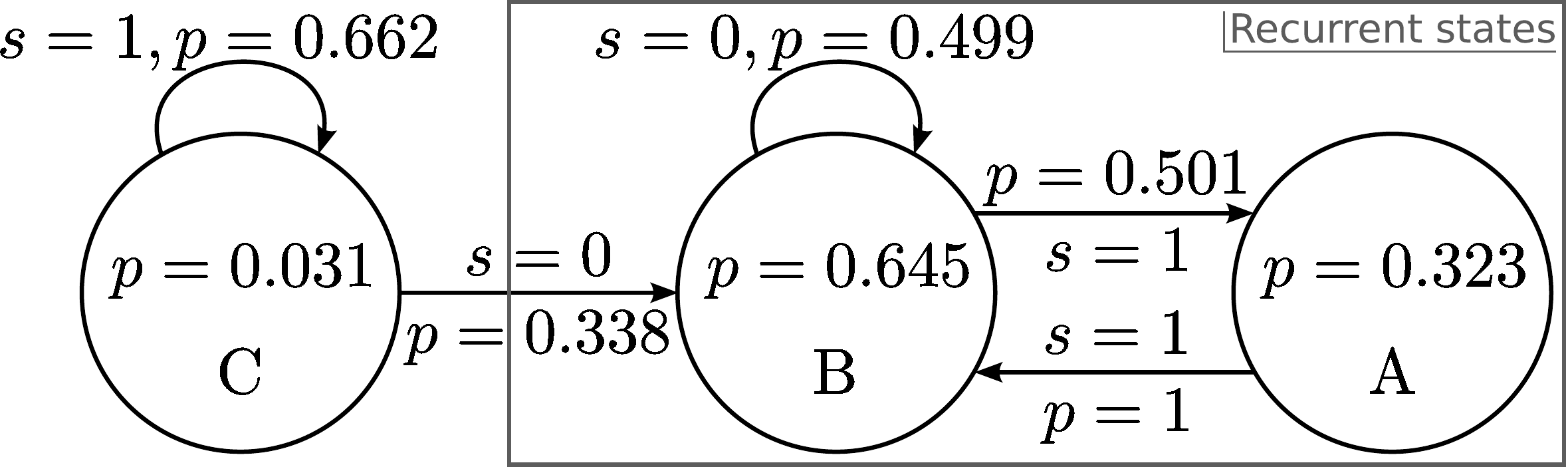}
\par\end{centering}

{\small Parameters: $10^{5}$ data points, using a past window size
of 10 points (random seed = 1).}{\small \par}

\caption{Reconstruction of the Even process.\label{fig:Even-process-rec}}
\end{figure}

The algorithm described in Section \ref{sec:Algo} is set up with
the following parameters:
\begin{itemize}
\item System configurations $x\in X$ are taken as the symbols in a sliding
window of size $L$ past data values. The predictions $z\in Z=\{0,1\}$
are the symbol in the series following this window, matching what
is used in CSSR.
\item Discrete distributions are built by monitoring $(x,z)$ pairs in the
training set of size $N$.
\item A Chi-Square test is used in order to match distributions, with $5\%$
accuracy.
\item The aggregative clustering algorithm described in Section \ref{sub:Clustering}
is applied, with a single pass.
\item Symbol constraints are available and implemented as described in Section
\ref{sub:Determinism-vs-CSSR} with a tolerance threshold $\lyxmathsym{θ}=0.95$.
\item We are not interested in this example in decisional states, so we
do not set a utility function.
\end{itemize}
The result of one reconstruction with a typical transient state, using
$N=10^{5}$ associations and a past window of 10 points, is shown
in Fig. \ref{fig:Even-process-rec}. The recurrent causal states of
this ε-machine correctly correspond to the definition of the Even
process. Close inspection of the data shows that the transient state
corresponds to strings formed of $10$ symbols $1$ in a row. Due
to the limit in window size the algorithm cannot distinguish whether
the last symbol $1$ was emitted from recurrent state A or B. Logically,
it observes that $\frac{1}{3}$ of the time the next symbol is a $0$
in the data set and $\frac{2}{3}$ of the time it is a $1$, matching
the proportions of the symbols in the data set: $p(s=1)=p(s=1|A)p(A)+p(s=1|B)p(B)$
as the process is really in either the state A or the state B.

The proposed algorithm classifies every single training point in a
causal state, hence creates transient states if necessary to match
data. Note that despite the Even process not being equivalent to any
finite state output HMM chain, the proposed algorithm reconstructs
it fairly well with a window size of 10.

In \cite[Fig. 4]{CSSR}, an experiment is conducted to study the behaviour
of the CSSR algorithm depending on the history size. The transposition
of this experiment is conducted in the current framework in order
to highlight the differences between both algorithm behaviours.

\begin{figure}[h]
\begin{centering}
\includegraphics[width=1\columnwidth]{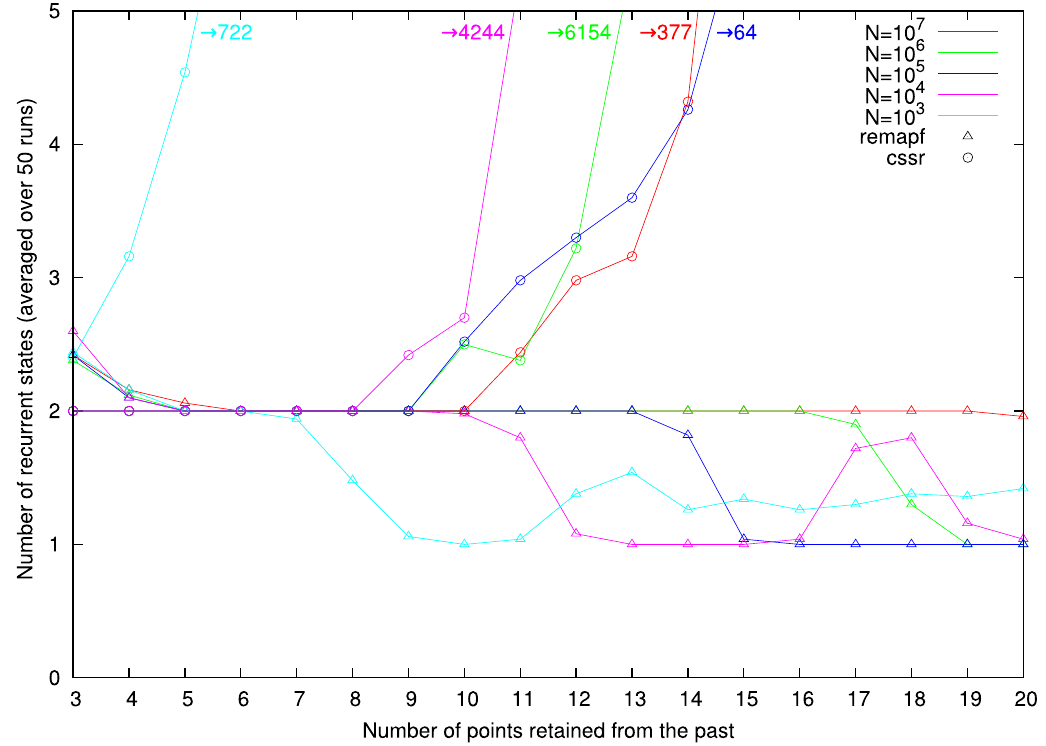}
\par\end{centering}

\caption{Number of recurrent states reconstructed from the Even process by
the REMAPF and CSSR algorithms.\label{fig:Even-process-nstates}}
\end{figure}

Figure \ref{fig:Even-process-nstates} shows the result of that experiment:
how many recurrent states are found on average (over 50 independent
trials) by the proposed algorithm and by CSSR, depending on the window
size. Results for $L=1$ and $L=2$ produce an incorrect transition
graph, there is not enough history to reliably determine the states,
and thus they are not presented in this figure.

Both algorithms reconstruct the states correctly when there is enough
data compared to the history length. Their behavior differs when the
number of observations is not enough to estimate the distributions
of histories correctly. This happens both when $N$ is too small (not
enough observations), and when $L$ is too large (too many possible
histories).

CSSR may split the states with each increase in the window size, whereas
the present algorithm clusters states using the whole window and symbol
constraints. When there is not enough data to estimate the distributions
properly CSSR over-splits the states. The current algorithm merges
them.

The proposed algorithm is also more robust, at least on this example.
Fig \ref{fig:Even-process-nstates} shows that for each data size
$N$ the REMAPF algorithm can correctly reconstruct the states for
a larger range of history size $L$ than CSSR. The larger $N$, and
the larger that range of histories.

Regarding the computation time, CSSR worst case complexity \cite{CSSR}
is $O(n_{e}^{2L+1})+O(N)$ where $n_{e}$ is the number of edges per
node in the reconstructed graph (hence $n_{e}$ is at most the alphabet
size $|A|$). On average CSSR is faster than its worst case, but still
exponential in history length. REMAPF is constant time with respect
to the history length. With the chosen clustering algorithm in this
section REMAPF is $O(KN)$ where $K$ is the number of estimated states.
For small history lengths CSSR is faster than REMAPF as $O(n_{e}^{2L+1})+O(N)\approx O(N)$
in that case, while REMAPF remains $O(KN)$. Fortunately REMAPF does
not oversplit the states, hence its computation time remains small
in the pathological cases, while CSSR computation time explodes. Table
\ref{tab:timings} shows the computation times averaged over 50 runs
for both algorithms and at each history size $L$ (thus 50 runs per
table cell, averaged), computed on a 2.2 GHz machine.

\begin{table*}[!]
\selectlanguage{english}%
\noindent %
\begin{tabular}{|l|l|l|l|l|l|l|l|l|l|l|l|l|l|l|l|l|l|l|l|}
\hline 
\multicolumn{1}{|l|}{{\footnotesize N}} & \begin{sideways}
\selectlanguage{british}%
{\footnotesize Algo}\selectlanguage{english}
\end{sideways} & \begin{sideways}
{\footnotesize L=3}
\end{sideways} & \begin{sideways}
{\footnotesize L=4}
\end{sideways} & \begin{sideways}
{\footnotesize L=5}
\end{sideways} & \begin{sideways}
{\footnotesize L=6}
\end{sideways} & \begin{sideways}
{\footnotesize L=7}
\end{sideways} & \begin{sideways}
{\footnotesize L=8}
\end{sideways} & \begin{sideways}
{\footnotesize L=9}
\end{sideways} & \begin{sideways}
{\footnotesize L=10}
\end{sideways} & \begin{sideways}
{\footnotesize L=11}
\end{sideways} & \begin{sideways}
{\footnotesize L=12}
\end{sideways} & \begin{sideways}
{\footnotesize L=13}
\end{sideways} & \begin{sideways}
{\footnotesize L=14}
\end{sideways} & \begin{sideways}
{\footnotesize L=15}
\end{sideways} & \begin{sideways}
{\footnotesize L=16}
\end{sideways} & \begin{sideways}
{\footnotesize L=17}
\end{sideways} & \begin{sideways}
{\footnotesize L=18}
\end{sideways} & \begin{sideways}
{\footnotesize L=19}
\end{sideways} & \begin{sideways}
{\footnotesize L=20}
\end{sideways}\tabularnewline
\hline 
\multirow{2}{*}{{\footnotesize 1e3}} & {\footnotesize R} & {\scriptsize 0.01} & {\scriptsize 0.01} & {\scriptsize 0.009} & {\scriptsize 0.01} & {\scriptsize 0.01} & {\scriptsize 0.01} & {\scriptsize 0.01} & {\scriptsize 0.01} & {\scriptsize 0.01} & {\scriptsize 0.01} & {\scriptsize 0.01} & {\scriptsize 0.02} & {\scriptsize 0.01} & {\scriptsize 0.01} & {\scriptsize 0.02} & {\scriptsize 0.01} & {\scriptsize 0.01} & {\scriptsize 0.01}\tabularnewline
\cline{2-20} 
 & {\footnotesize C} & {\scriptsize 0.01} & {\scriptsize 0.01} & {\scriptsize 0.009} & {\scriptsize 0.01} & {\scriptsize 0.01} & {\scriptsize 0.01} & {\scriptsize 0.02} & {\scriptsize 0.02} & {\scriptsize 0.04} & {\scriptsize 0.08} & {\scriptsize 0.14} & {\scriptsize 0.20} & {\scriptsize 0.29} & {\scriptsize 0.34} & {\scriptsize 0.40} & {\scriptsize 0.49} & {\scriptsize 0.58} & {\scriptsize 0.68}\tabularnewline
\cline{2-20} 
\multirow{2}{*}{{\footnotesize 1e4}} & {\footnotesize R} & {\scriptsize 0.02} & {\scriptsize 0.03} & {\scriptsize 0.02} & {\scriptsize 0.03} & {\scriptsize 0.02} & {\scriptsize 0.03} & {\scriptsize 0.02} & {\scriptsize 0.03} & {\scriptsize 0.03} & {\scriptsize 0.03} & {\scriptsize 0.04} & {\scriptsize 0.04} & {\scriptsize 0.06} & {\scriptsize 0.1} & {\scriptsize 0.14} & {\scriptsize 0.21} & {\scriptsize 0.26} & {\scriptsize 0.27}\tabularnewline
\cline{2-20} 
 & {\footnotesize C} & {\scriptsize 0.01} & {\scriptsize 0.01} & {\scriptsize 0.01} & {\scriptsize 0.02} & {\scriptsize 0.01} & {\scriptsize 0.02} & {\scriptsize 0.02} & {\scriptsize 0.02} & {\scriptsize 0.03} & {\scriptsize 0.05} & {\scriptsize 0.13} & {\scriptsize 0.48} & {\scriptsize 1.52} & {\scriptsize 4.3} & {\scriptsize 10.5} & {\scriptsize 22.5} & {\scriptsize 41.6} & {\scriptsize 72.1}\tabularnewline
\cline{2-20} 
\multirow{2}{*}{{\footnotesize 1e5}} & {\footnotesize R} & {\scriptsize 0.16} & {\scriptsize 0.17} & {\scriptsize 0.19} & {\scriptsize 0.17} & {\scriptsize 0.19} & {\scriptsize 0.2} & {\scriptsize 0.19} & {\scriptsize 0.18} & {\scriptsize 0.21} & {\scriptsize 0.21} & {\scriptsize 0.20} & {\scriptsize 0.21} & {\scriptsize 0.29} & {\scriptsize 0.36} & {\scriptsize 0.55} & {\scriptsize 1.02} & {\scriptsize 2.17} & {\scriptsize 5.03}\tabularnewline
\cline{2-20} 
 & {\footnotesize C} & {\scriptsize 0.04} & {\scriptsize 0.04} & {\scriptsize 0.04} & {\scriptsize 0.05} & {\scriptsize 0.05} & {\scriptsize 0.05} & {\scriptsize 0.05} & {\scriptsize 0.06} & {\scriptsize 0.06} & {\scriptsize 0.08} & {\scriptsize 0.11} & {\scriptsize 0.20} & {\scriptsize 0.44} & {\scriptsize 1.7} & {\scriptsize 7.82} & {\scriptsize 28} & {\scriptsize 120} & {\scriptsize 512}\tabularnewline
\cline{2-20} 
\multirow{2}{*}{{\footnotesize 1e6}} & {\footnotesize R} & {\scriptsize 1.5} & {\scriptsize 1.49} & {\scriptsize 1.5} & {\scriptsize 1.54} & {\scriptsize 1.53} & {\scriptsize 1.54} & {\scriptsize 1.55} & {\scriptsize 1.62} & {\scriptsize 1.6} & {\scriptsize 1.64} & {\scriptsize 1.73} & {\scriptsize 1.74} & {\scriptsize 1.94} & {\scriptsize 1.95} & {\scriptsize 2.24} & {\scriptsize 2.95} & {\scriptsize 4.74} & {\scriptsize 14.4}\tabularnewline
\cline{2-20} 
 & {\footnotesize C} & {\scriptsize 0.3} & {\scriptsize 0.31} & {\scriptsize 0.33} & {\scriptsize 0.35} & {\scriptsize 0.36} & {\scriptsize 0.38} & {\scriptsize 0.39} & {\scriptsize 0.41} & {\scriptsize 0.44} & {\scriptsize 0.47} & {\scriptsize 0.53} & {\scriptsize 0.64} & {\scriptsize 0.87} & {\scriptsize 2.1} & {\scriptsize 7.45} & {\scriptsize 23.4} & {\scriptsize 69.5} & {\scriptsize 217}\tabularnewline
\cline{2-20} 
\multirow{2}{*}{{\footnotesize 1e7}} & {\footnotesize R} & {\scriptsize 14.7} & {\scriptsize 14.7} & {\scriptsize 14.9} & {\scriptsize 15.1} & {\scriptsize 15.1} & {\scriptsize 15.2} & {\scriptsize 15.3} & {\scriptsize 15.8} & {\scriptsize 15.8} & {\scriptsize 16.2} & {\scriptsize 16.4} & {\scriptsize 17} & {\scriptsize 17.6} & {\scriptsize 18.1} & {\scriptsize 19.2} & {\scriptsize 22.4} & {\scriptsize 26.9} & {\scriptsize 40.7}\tabularnewline
\cline{2-20} 
 & {\footnotesize C} & {\scriptsize 2.89} & {\scriptsize 3.11} & {\scriptsize 3.25} & {\scriptsize 3.41} & {\scriptsize 3.54} & {\scriptsize 3.6} & {\scriptsize 3.74} & {\scriptsize 3.92} & {\scriptsize 4.1} & {\scriptsize 4.32} & {\scriptsize 4.6} & {\scriptsize 4.94} & {\scriptsize 5.45} & {\scriptsize 6.97} & {\scriptsize 12.7} & {\scriptsize 29.2} & {\scriptsize 74} & {\scriptsize 200}\tabularnewline
\hline 
\end{tabular}

\selectlanguage{british}%
\caption{\label{tab:timings}Timings (seconds) for the experiments in Fig.\ref{fig:Even-process-nstates}}
\end{table*}

\subsection{\label{sub:Cellular-automaton}Cellular automaton}

Another test case where causal states were applied is the detection
of moving particles in cellular automata, and their interactions \cite{filters_ca_stat_comp}.
The introduction of a utility function in this context provides a
simple yet effective way to demonstrate the concepts presented in
this document. The next section considers the usage of decisional
states in a larger-scale application.

\begin{figure}
\begin{centering}
\includegraphics[width=1\columnwidth]{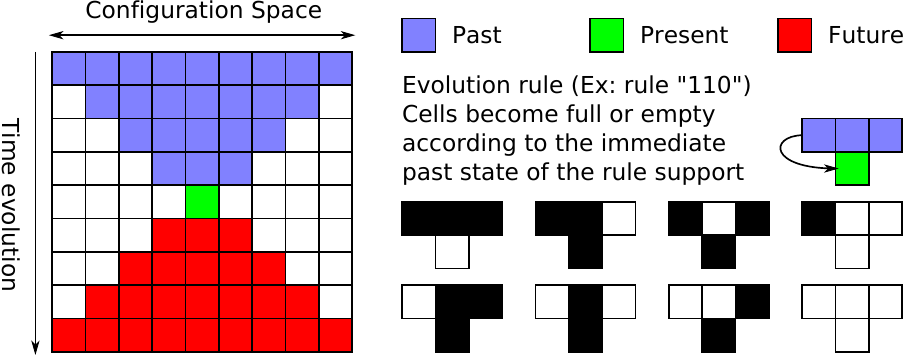}
\par\end{centering}

\caption{Elementary cellular automaton.\label{fig:Elementary-cellular-automaton.}}
\end{figure}

Figure \ref{fig:Elementary-cellular-automaton.} shows the evolution
rule and configuration of an elementary one-dimensional automaton.
Each row of the regular grid contains the system state at a given
time. An evolution rule dictates how the cell binary states evolve
at each time step. The numbering scheme {}``110'' refers to an exhaustive
listing of all possible elementary cellular automata rules \cite{wolframAutomatonRules}.
The evolution rule has a support, 3 cells in figure \ref{fig:Elementary-cellular-automaton.},
from which the next cell configuration is deduced. Propagation of
this support in time defines {}``light-cones'' according to the
terminology of Appendix A, within the implementation constraint of
a limited depth.

In this context the data set $X$ is the space of all past light-cones
(in blue on Figure \ref{fig:Elementary-cellular-automaton.}). From
the current system state we would like to predict the future of the
system, so $Z$ is the space of all future light-cones (in red on
Figure \ref{fig:Elementary-cellular-automaton.}). Even though the
cellular automaton is completely deterministic, the state of cells
in the future cone depend on information which is outside the past
cone, so we observe a distribution of different futures for each past
(see Section \ref{sub:Causality_and_predictions}). With cyclic boundary
conditions and a fixed evolution rule for the whole automaton, all
cells have exactly the same distribution so we can aggregate the observations
across all cells.

The utility function is chosen by the user according to the application
needs. Here we chose to define the utility of a prediction as the
number of correctly predicted cell states in the future cone. Hence
the utility takes in this example integer values between $0$ and
the maximum $d^{2}-1$ where $d$ if the future cone depth (we could
also have used a proportion between $0$ and $1$).

Given the discrete nature and relatively small search space of the
problem, the algorithm described in Section \ref{sub:AlgoSetup} is
setup with:
\begin{itemize}
\item A simple discrete probability density estimator based on $(\boldsymbol{x},\boldsymbol{z})$
observation counts: $\hat{p}(\boldsymbol{z}|\boldsymbol{x})=\frac{count(\boldsymbol{x},\boldsymbol{z})}{count(\boldsymbol{x})}$.
\item An exhaustive integrator weighting the utility of all possible future
cones by their probability: $\hat{\mathbb{U}}(y|x)=\sum_{z\in Z}\hat{p}(z|x)U(y,z)$.
In practice unobserved $z$ values would induce a null contribution
so the summing occurs only on observed $\boldsymbol{z}$.
\item An exhaustive search optimiser, computing $\hat{\mathbb{U}}(y|x)$
for all possible $y\in Z$. The maximum utility value as well as the
set $Y(x)=\mathrm{Argmax}_{y\in Z}\hat{\mathbb{U}}(y|x)$ of best
predictions are maintained during the search.
\item The connected component, single-link hierarchical clustering algorithm
described in Section \ref{sub:Clustering}, with exact match predicates.
\end{itemize}
\begin{figure}
\begin{centering}
\includegraphics[width=1\columnwidth]{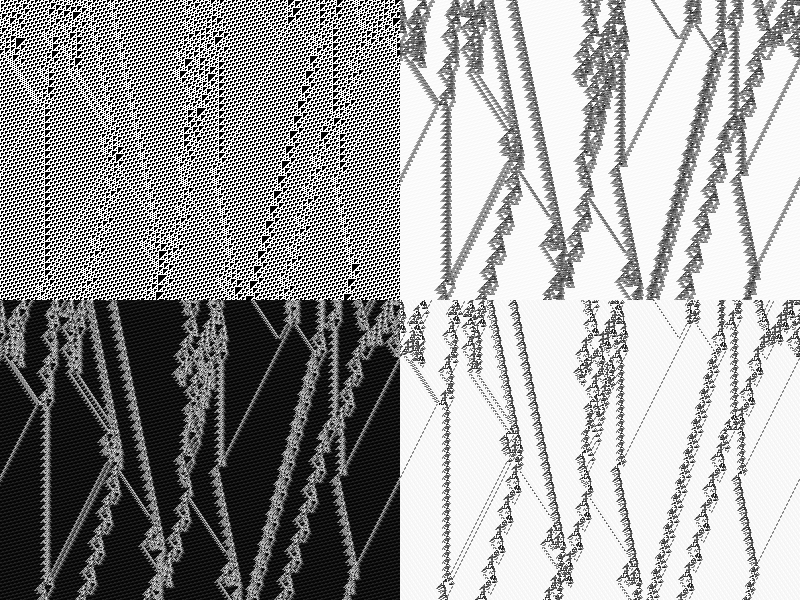}
\par\end{centering}

\noindent {\small Top-left: Cellular automaton cell values (1 is white,
0 is black).}{\small \par}

\noindent {\small Top-right: Local statistical complexity field (difficulty
to get the future distribution) scaled between white (minimal value)
and black (maximal complexity).}{\small \par}

\noindent {\small Bottom-left: Local expected utility field (not the
complexity, the value of the expected utility itself) scaled similarly.}{\small \par}

\noindent {\small Bottom-right: Local iso-prediction complexity field
(difficulty to get an optimal prediction) scaled similarly.}{\small \par}

\noindent {\small Parameters: Past depth 4, future depth 3, 400 cells,
300 steps, 100 initial transient dropped.}{\small \par}

\caption{Raw cells and complexity fields of a cellular automaton. \label{fig:rule110results}}
\end{figure}
Figure \ref{fig:rule110results} shows the results of the experiment
for the evolution rule introduced in Fig. \ref{fig:Elementary-cellular-automaton.}.
Figure \ref{fig:rule110results} is directly comparable with \cite[Fig. 3]{filters_ca_stat_comp},
where another algorithm was used to estimate the ε-machine. The raw
cellular automaton field is the direct application of the rule depicted
in Fig. \ref{fig:Elementary-cellular-automaton.}. The statistical
and iso-prediction complexity fields are mapped to a grey scale range
where white represent their respective minimum complexity value and
black their respective maximum. The expected utility field is also
shown in \ref{fig:rule110results}.

The causal states sub-cluster the iso-predictive states: we observe
alternative versions of the particles. The utility is based on the
number of correctly predicted cells, irrespectively of their position
in the future cone. Information that is irrelevant to this utility
function is masked out in the iso-prediction field, whereas it was
present in the statistical complexity field. 

Some information was lost. But if all the user cares is encoded in
the utility function, that information was noise and clarity was gained
in the result. In extension to \cite{filters_ca_stat_comp}, we have
defined a new family of automatic filters based on utility functions.

\subsection{\label{sub:Image-segmentation}Image filtering and edge detection}

The idea of this section is to extend the cellular automaton example
for filtering images. We make the hypothesis that edges correspond
to zones where the prediction difficulty is greatest. This differs
from other common definitions, like a high luminance gradient magnitude.
The definition of an edge is not the topic of this article. This section's
goal is to show how the concepts introduced in this document might
be used on a concrete non-temporal data example.

Following the construction in \cite{AwateThesis}, the data space
$X$ is defined as the neighbourhoods of image pixels $z\in Z$. The
prediction problem is to find the value of $z$ from the neighbourhood.
\begin{figure}
\begin{centering}
\includegraphics[width=1\columnwidth]{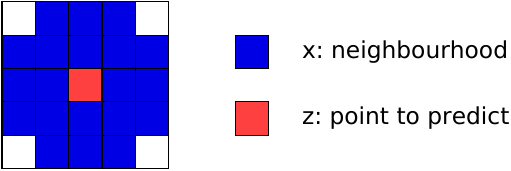}
\par\end{centering}

\caption{Neighbourhood for image filtering.\label{fig:Neighbourhood}}
\end{figure}
Figure \ref{fig:Neighbourhood} shows how the neighbourhood is defined
in this experiment: up to two pixels in each direction except corners.
Larger or smaller neighbourhoods were tested: smaller regions make
the prediction more difficult, larger regions lead to thicker detected
edges. Similarly defining $z$ as a centre block instead of a single
pixel was also tried, with similar observed effects (precision and
edge size). For a 8-bits grey image the data space is thus $X=\left[0\ldots255\right]^{20}$
and the prediction space $Z=\left[0\ldots255\right]$.

The data space $X$ is thus considerably larger than in the previous
examples. Fortunately, unlike the cellular automaton case where a
difference in $x$ can lead to completely different outcomes, usually
images are not significantly altered when pixel values differ by a
small amount. In the present context we exploit this nearby consistency
in $X$ and $Z$ in order to apply kernel density estimators. These
are more reliable than the simple count-based estimator used in the
previous section, especially since $X$ has a higher dimension.

The approach of considering the image as the limit distribution of
a Markov Random Field \cite{AwateThesis} is applied to this example.
Concretely, this amounts to estimating the probability densities in
the joint space $X\times Z$ and inferring the conditional distributions
by integration of $\hat{P}(X)$, as described in Section \ref{sub:Kernel-density-estimation}.

The prediction space $Z$ can be run through exhaustively in the case
considered here: only one centre grey pixel. Sub-sampling for numerical
integration is thus not necessary, and we set $S=Z$.

The utility of a prediction $y$ when the true value $z$ happens
is defined as $U(y,z)=-\max\left(0,\left|y-z\right|-\tau\right)$.
In other words small prediction discrepancies up to $\tau$ are accepted
at no cost, reflecting the fact the image is not significantly altered
by small variations in pixel grey levels. Then the utility decreases
(the loss increases) with each grey level difference between the predicted
and the true value.

\begin{figure*}
\begin{centering}
\includegraphics[width=1\textwidth]{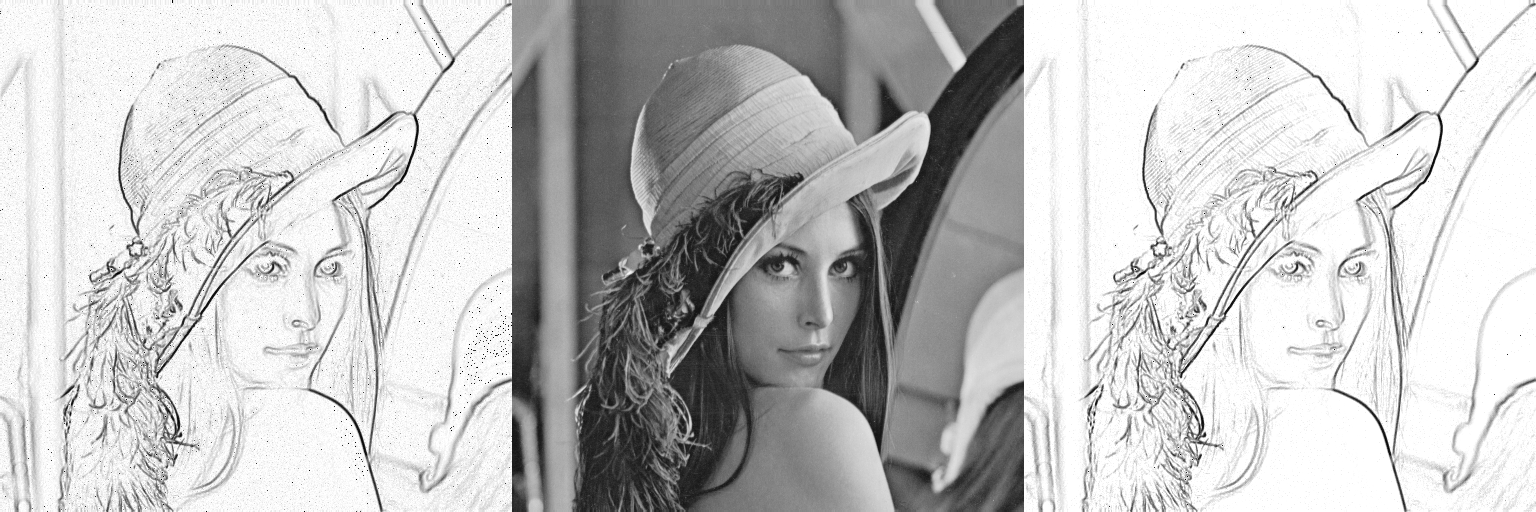}
\par\end{centering}

Left: Filter obtained with local statistical complexity ranks (see
main text)

Middle: Original Lenna Image

Right: Filter obtained with local iso-prediction complexity ranks.
The background noise is eliminated but the edges are preserved.

\caption{Proposed Image filter (top left) and some variants.\label{fig:imageFilterExample}}
\end{figure*}

A pre-precessing is performed. For each region $x$ the minimal grey
level is computed. It is then subtracted from both $x$ and $z$ without
loss of genericity (the original grey level for a prediction $z$
can be reconstructed by adding back the value shift defined on $x$).
A post-processing is performed: ordering the states by their complexity
values and plotting their ranks with respect to that ordering on a
grey scale. Fig. \ref{fig:imageFilterExample} shows the result of
this experiment. 

The effect of applying a utility function is apparent on the right
part of Fig. \ref{fig:imageFilterExample}. The noise inherent present
in the image filtered by causal states may be attributed to errors
in the density estimation and clustering technique: some extra states
that appear only once, thus with high complexity value (black dots).
When applying a utility function these extra states are merged into
coarser states with the same properties with respect to that utility
function. The number of black dots thus decreases. The utility function
also manifests itself in the reduction of the background noise in
the image. Since a tolerance is given the small differences in grey
levels are ignored (up to $\tau=7$ in this example at no cost). The
flat zones in the picture are thus more uniformely white (ex: the
shoulder at the middle-bottom of the picture).

The filter defined by local complexity values has several characteristics
that differ from other more traditional image processing techniques:
\begin{itemize}
\item The filter is defined globally: features are defined on the whole
picture. For example the flat zones in the original image are detected
as having low complexity. When the filter is applied locally this
information is taken into account. The extreme example would be the
cellular automaton background in Figure \ref{fig:rule110results}:
many edges would be detected for each small triangle pattern using
a simple gray level gradient based filtering, but the proposed filter
assigns a low value for these repetitive patterns.
\item Fine details are similarly considered statistically on the whole picture.
In the bottom-left region of the picture, the filaments attached to
the hat are detected as single units: each light-dark transition has
its own complexity, low values are whitened out by the ranking. This
also differs from gray level gradient filtering techniques that are
prone to emit an edge on each side of the filament.
\item Making global statistics and clustering probability distributions
comes with a computational cost. The time consuming task is to build
the statistics: it took 37 minutes on a 2.2Ghz quad-core CPU to build
the probability distributions for the Fig. \ref{fig:imageFilterExample}
with a kernel width of $1.5$ units in the joint space $X\times Z$.
By comparison clustering the causal states with 3 passes of the linear
clustering took 57 seconds, and applying the utility took 100 ms.
\end{itemize}
It would be interesting to combine the complexity-based filters to
other well-established edge detection, segmentation or noise removal
algorithms. The goal of this article is to introduce the decisional
states and their applications in different contexts. The proposed
edge detection is only a demonstration of how the main concept might
be used in practice, this framework may possibly be adapted for generic
feature detection with a variant setup (especially alternative utility
functions and larger support for the pixels in the $X$ space).

\section{Conclusion}

The decisional states notion was introduced: the internal states of
a system that lead to the same decision, given a user-defined utility
function. Compared to alternative approaches in the domain \cite{MDP_60,UtileDistinctionHMM},
here the utility function is defined on the space of predictions:
$U(y,z)$ quantifies what gain/loss is incurred when $y$ is predicted
while $z$ happens. This makes the present work suited for applications
like time series processing and detection of anomalous/more complex
zones in a system, while less suited for reinforcement learning \cite{RRL_overview}.

The natural framework for applying the utility function is the ε-machine
\cite{CrutchfieldStatComp}, which is an edge-emitting Markovian graph
model with higher genericity than the usual state-output Markov chains
\cite{DMarkovMachine}. In this context the ε-machine corresponds
to the internal structure of the system, irrespectively of any user-defined
utility. The decisional states are built on top of this internal structure
in a way that reflects the external knowledge (encoded in the utility
function) brought in the system.

Coming with the decisional states are definitions of complexity measures
on the system. It is possible to quantify precisely, in number of
bits, the difficulty of making an optimal prediction in terms of the
chosen utility. Another consequence is a way to identify events that
provoke a change of decision, represented as transitions in a state
diagram, assuming decisions are based on the expected utility. The
decisional states were exemplified mathematically on analytically
tractable examples, and numerically on practical problems like image
filtering. A separate article \cite{eegdstates} shows how to use
this notion on large data sets of real EEG signals.

A new algorithm was introduced for computing an ε-machine from observed
data and for computing the newly introduced decisional states on top
of it. This algorithm is very adaptable to specific application needs,
including for temporal and spatial data, using a symbolic representation
or not, as demonstrated by the examples in the previous sections.
A reference implementation is provided, see the Appendix. It is available
as free-libre software.

\section*{Appendix: Web information}

The latest experimental version of the code as well as previous versions
are available at the source repository at http://source.numerimoire.net/decisional\_states

The code is highly templatised and the classes might be directly included
into a user project without having to link to an external library.
The code is available as free-libre software (GNU LGPL v2.1 or more
recent) and contributions are welcome.

\bibliographystyle{abbrv}
\bibliography{decisional_states}

\end{document}